\documentclass[]{bytedance_seed}

\usepackage[toc,page,header]{appendix}

\usepackage{minitoc}

\usepackage{microtype}
\usepackage{hyperref}
\usepackage{url}
\usepackage{booktabs}

\usepackage{graphicx}

\usepackage{lineno}

\usepackage{algorithm}
\usepackage{algpseudocode}

\usepackage{makecell}

\usepackage{wrapfig}
\usepackage{overpic}
\usepackage{float}
\usepackage{rotating}

\usepackage{float}
\usepackage{algorithmicx}
\restylefloat{algorithm}  % 允许 algorithm 换页

\usepackage{amsmath}
\usepackage{amssymb}
\usepackage{mathtools}
\usepackage{amsthm}
\usepackage{listings}

\usepackage{caption}
\usepackage{etoolbox}
\makeatletter
\AfterEndEnvironment{algorithm}{\let\@algcomment\relax}
\AtEndEnvironment{algorithm}{\kern2pt\hrule\relax\vskip3pt\@algcomment}
\let\@algcomment\relax
\newcommand\algcomment[1]{\def\@algcomment{\footnotesize#1}}
\renewcommand\fs@ruled{\def\@fs@cfont{\bfseries}\let\@fs@capt\floatc@ruled
  \def\@fs@pre{\hrule height.8pt depth0pt \kern2pt}%
  \def\@fs@post{}%
  \def\@fs@mid{\kern2pt\hrule\kern2pt}%
  \let\@fs@iftopcapt\iftrue}
\makeatother

\definecolor{darkblue}{rgb}{0, 0, 0.5}
\hypersetup{colorlinks=true, citecolor=darkblue, linkcolor=darkblue, urlcolor=darkblue}

\title{Frac-Connections: Fractional Extension of Hyper-Connections}

\author[1, \dagger]{Defa Zhu}
\author[1]{Hongzhi Huang}
\author[1]{Jundong Zhou}
\author[1]{Zihao Huang}
\author[1]{Yutao Zeng}
\author[1]{Banggu Wu}
\author[1, \dagger]{Qiyang Min}
\author[1]{Xun Zhou}

\affiliation[1]{ByteDance Seed}

\contribution[\dagger]{Corresponding authors}

\abstract{
Residual connections are central to modern deep learning architectures, enabling the training of very deep networks by mitigating gradient vanishing. Hyper-Connections recently generalized residual connections by introducing multiple connection strengths at different depths, thereby addressing the seesaw effect between gradient vanishing and representation collapse. However, Hyper-Connections increase memory access costs by expanding the width of hidden states. In this paper, we propose Frac-Connections, a novel approach that divides hidden states into multiple parts rather than expanding their width. Frac-Connections retain partial benefits of Hyper-Connections while reducing memory consumption. To validate their effectiveness, we conduct large-scale experiments on language tasks, with the largest being a 7B MoE model trained on up to 3T tokens, demonstrating that Frac-Connections significantly outperform residual connections.
}

\date{\today}
\correspondence{Defa Zhu at \email{zhudefa@bytedance.com}, Qiyang Min at \email{qiyangming@bytedance.com}}

\begin{document}
\maketitle

%不需要目录就注释掉 注意目录不要和第一页放在一块 要有\newpage
%\newpage
%\tableofcontents
%\newpage

\section{Introduction}
\begin{figure}[h]
    \begin{center}
    \includegraphics[width=\columnwidth]{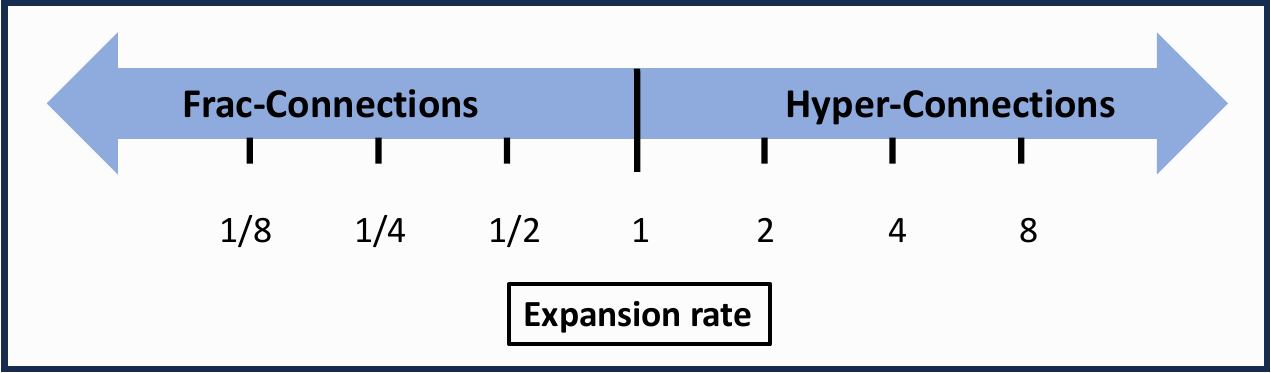}
    \end{center}
  \caption{Comparison of Frac-Connections and Hyper-Connections based on their expansion rates. Frac-Connections correspond to \(n \leq 1\), while Hyper-Connections are defined by \(n \geq 1\). The two connection types become identical when the expansion rate is \(n = 1\).}
    \label{fig:fcvshc}
\end{figure}

Residual connections \citep{he2016deep} have revolutionized deep learning by facilitating the effective training of very deep networks. These connections mitigate gradient vanishing and are fundamental to architectures such as transformers and convolutional neural networks (CNNs). However, residual connections suffer from a trade-off between gradient vanishing and representation collapse, where the features of adjacent layers become excessively similar, particularly in very deep models~\citep{xie2023residual,liu2020understanding,zhu2024hyper}.

\citet{zhu2024hyper} introduce Hyper-Connections, an expansion of the dimension of hidden state and learnable depth and width connections, to address this issue. While effective, Hyper-Connections increase memory access by expanding the hidden states' width. This raises the question: \textit{Can we enjoy the benefits of Hyper-Connections without increasing memory access?}

\begin{figure}[h]
    \centering
    \begin{overpic}[abs,unit=1mm,scale=0.2,width=0.4\textwidth]{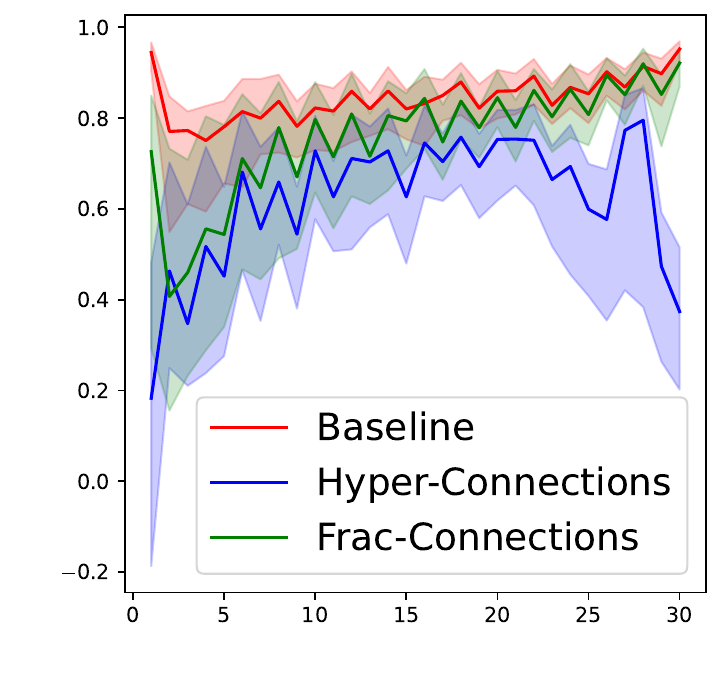}
        \put(0,17){\small\begin{turn}{90} $\texttt{cos}(\mathbf{h}_0^i, \mathbf{h}_0^{i+1})$ \end{turn}}
        \put(20,0){\small Layer Index $i$}
    \end{overpic}
    \caption{Cosine similarity between the input of the current and the previous layers for the \texttt{OLMoE-7B} models. The curve represents the median of similarity, while the shaded area indicates the range between the 5th and 95th percentiles.}
    \label{fig:compare_hidden_sim_curve}
\end{figure}
To this end, we propose Frac-Connections (FC), a novel method that partitions the hidden states into multiple fractions rather than duplicating them and increasing their width. This approach extends the expansion rate $n$ of Hyper-Connections (HC) to the fractional domain. In particular, when $n = 1$, Frac-Connections and Hyper-Connections are equivalent, as illustrated in Fig.~\ref{fig:fcvshc}. This reduces memory usage while preserving the ability to model multiple connection strengths. As shown in Fig.\ref{fig:compare_hidden_sim_curve}, the similarity between adjacent hidden states in FC lies between that of HC and baseline (Pre-Norm), indicating that their representational capacity follows the order: HC $>$ FC $>$ Pre-Norm.

To further validate the effectiveness of Frac-Connections, we conduct extensive experiments on large language models (LLMs), including both dense and Mixture-of-Experts (MoE)~\citep{shazeer2017sparsely} architectures. Our results demonstrate that Frac-Connections significantly improve training stability and enhance downstream task performance across a wide range of natural language processing benchmarks. We believe that the simplicity, scalability, and efficiency of Frac-Connections will enable their widespread adoption across various domains in machine learning, providing a robust foundation for building the next generation of dense and sparse deep learning models.
\section{Related Work}
\textbf{Transformers} \citep{vaswani2017attention,dosovitskiy2020image,huang2024ultra,huang2025over,wang2025scale} have revolutionized deep learning, particularly in natural language processing and computer vision. They rely on self-attention mechanisms to capture long-range dependencies and have become the foundation of large-scale models such as BERT \citep{devlin2019bert} and GPT \citep{brown2020language}. A key component of Transformers is residual connections \citep{he2016deep}, which aid training but may also limit model expressiveness \citep{zhu2024hyper}. Our work focuses on replacing these residual connections to further enhance Transformer performance.

\textbf{Residual Connections and Their Limitations.}  
Residual connections \citep{he2016deep} have been a key component in modern deep networks, enabling the training of very deep architectures by mitigating the gradient vanishing problem. They are widely used in networks such as CNNs\citep{krizhevsky2012imagenet} and Transformers \citep{vaswani2017attention}. However, despite their effectiveness, residual connections introduce a fundamental trade-off between gradient propagation and representation collapse \citep{zhu2024hyper}, which can degrade performance in extremely deep models. ResiDual \citep{xie2023residual} addresses this issue by adopting a dual-stream design with parallel PreNorm and PostNorm structures, while Hyper-Connections use a weighted multi-stream design to significantly improve performance. While this improves performance, the multi-stream approach increases memory consumption. Our Frac-Connections build upon this design by reducing the hidden size of each stream, retaining the benefits of Hyper-Connections without increasing memory usage.

\textbf{Fractal Design. }
FractalNet \citep{larsson2016fractalnet} proposes partitioning the hidden states into multiple segments, each processed by networks of varying depths, enabling the training of extremely deep neural networks. Frac-Connections share a similar design principle; however, instead of assigning each partition to a different depth, we associate them with different connection weights.

\section{Preliminaries}
\begin{figure*}[t]
    \begin{center}
    \includegraphics[width=1\textwidth]{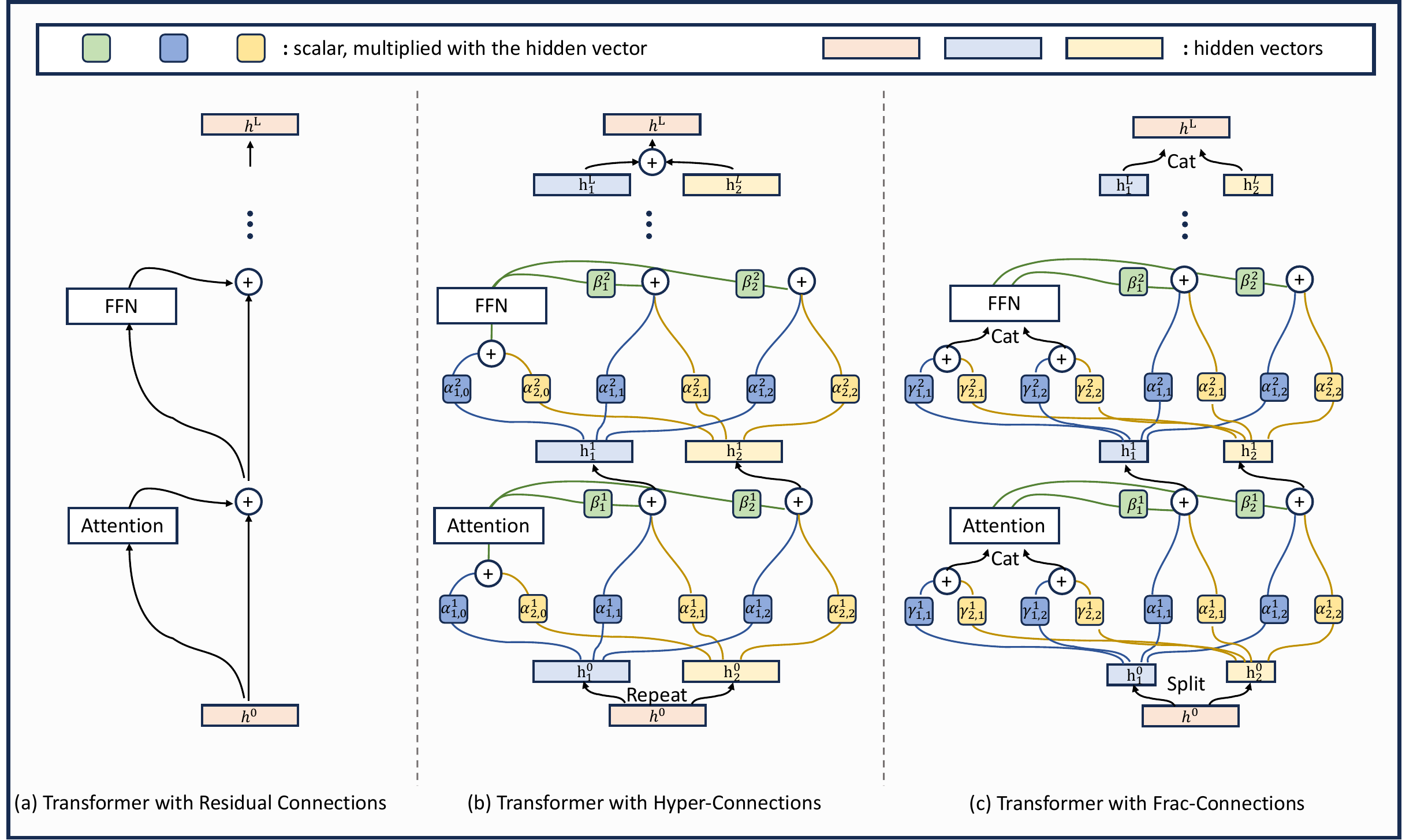}
    \end{center}
  \caption{\textbf{Figure 2. Frac-connections (FC) with an expansion rate of $n = 1/2$.} 
(a) Residual connections. 
(b) Hyper-connections: $\beta_1$, $\beta_2$, $\alpha_{0,0}$, $\alpha_{0,1}$, $\alpha_{1,0}$, $\alpha_{1,1}$, $\alpha_{2,1}$, and $\alpha_{2,2}$ are learnable scalars or scalars predicted by the network, depending on the specific HC version. 
(c) Frac-connections: Frac-connections split the hidden representations into smaller fractions and process each fraction independently. The scalars $\gamma_{1,2}$, $\gamma_{2,1}$, and $\gamma_{2,2}$ are either learnable or predicted by the network, similar to hyper-connections. These fractions are concatenated (denoted as $\text{Cat}$) after processing, followed by integration into the main network pipeline.}
    \label{fig:fc_overview}
\end{figure*}

Hyper-Connections (HC) enhance the representation of hidden states in neural networks by introducing a \textbf{hyper hidden matrix}. Given the initial input $\mathbf{h}^0 \in \mathbb{R}^d$, it is replicated $n$ times to construct the initial hyper hidden matrix:

\begin{equation}
\mathbf{H}^0 = 
\begin{pmatrix}
\mathbf{h}^0 & \mathbf{h}^0 & \dots & \mathbf{h}^0
\end{pmatrix}^\intercal 
\in \mathbb{R}^{n \times d},
\end{equation}

where $n$ is the \textbf{expansion rate}. At the $k$-th layer, the input is the hyper hidden matrix from the previous layer, denoted as $\mathbf{H}^{k-1}$:

\begin{equation}
\mathbf{H}^{k-1} = 
\begin{pmatrix}
\mathbf{h}_1^{k-1} & \mathbf{h}_2^{k-1} & \dots & \mathbf{h}_n^{k-1}
\end{pmatrix}^\intercal \in \mathbb{R}^{n \times d}.
\end{equation}

The final hidden vectors are aggregated using \textbf{sum pooling}, which reduces the hyper hidden matrix back to a single vector.

The hyper-connections are modeled by a matrix $\mathcal{HC} \in \mathbb{R}^{(n+1) \times (n+1)}$, which defines the connection weights across different components:

\begin{equation}
\mathcal{HC}^k = 
\begin{pmatrix}
\mathbf{0}_{1 \times 1} & \mathbf{B}^k \\
\mathbf{A_m}^k & \mathbf{A_r}^k
\end{pmatrix},
\end{equation}

where $\mathbf{B}^k$, $\mathbf{A_m}^k$, and $\mathbf{A_r}^k$ are submatrices that define the connections within and between layers.

For a given network layer $\mathcal{T}^{k}$, which integrates components such as self-attention and feed-forward networks, the output $\mathbf{H}^{k+1}$ of the hyper-connections can be expressed as:

\begin{equation}
\mathbf{H}^k = {\mathbf{B}^{k}}^\intercal (\mathcal{T}^k (\mathbf{h}^{k-1}_0))^\intercal + {\mathbf{A_r}^k}^\intercal \mathbf{H}^{k-1},
\end{equation}

where $\mathbf{h}^k_0$ is computed as the weighted sum of the hyper hidden matrix using $\mathbf{A_m}^{k}$:

\begin{equation}
{\mathbf{h}^{k-1}_0}^\intercal = \mathbf{A_m}^\intercal \mathbf{H}^{k-1}.
\end{equation}

These matrices capture the relationships across both the depth and width dimensions of the network and are visualized in Fig.~\ref{fig:fc_overview}.

To further improve flexibility of the connections, \textbf{Dynamic Hyper-Connections (DHC)} extend this framework by making the weights input-dependent. Instead of using fixed parameters, the connection weights are dynamically predicted based on the input hidden vector $\mathbf{H}^{k}$. This adaptive mechanism improves its ability to represent complex relationships. The advantages of DHC are particularly evident in tasks such as language modeling.
\section{Method}
\subsection{Overview of Frac-Connections}
The purpose of introducing \textit{frac-connections} is to address the seesaw problem in residual connections while retaining the flexibility of constructing connection strengths, without incurring the additional memory overhead of splitting hidden states into $n$ parts as in \textit{hyper-connections}. This is achieved by generalizing the expansion rate to fractional values. When $n = 1$, frac-connections are equivalent to hyper-connections. For \( 0 < n < 1 \), frac-connections can be viewed as a fractional variant of hyper-connections that divides the hidden states into \( m = 1/n \) parts instead of replicating them \( n \) times, where \( m \) (referred to as the \textbf{frac-rate}) represents the number of partitions.

Let $\mathbf{h} \in \mathbb{R}^d$ represent the hidden state of a layer. Instead of replicating $\mathbf{h}$ into $n$ copies as in hyper-connections, frac-connections split $\mathbf{h}$ into $m = 1/n$ parts:
\begin{equation}
\mathbf{H} = \begin{pmatrix} \mathbf{h}_1 & \mathbf{h}_2 & \dots & \mathbf{h}_m \end{pmatrix}^\intercal = \texttt{Reshape}(\mathbf{h},(m,d/m)),
\end{equation}
where $\mathbf{h}_i \in \mathbb{R}^{d/m}$ for $i = 1, 2, \dots, m$.

The Frac-Connections (FC) can be represented by a matrix $\mathcal{FC}$, where each element defines the connection weight. The matrix is structured as follows:

\begin{align}
\label{eq:shc}
\mathcal{FC} &=
\begin{pmatrix}
\mathbf{0}_{1\times m} & \mathbf{B} \\
\mathbf{Y} & \mathbf{A}
\end{pmatrix} \nonumber \in \mathbb{R}^{(m+1) \times (2\times m)} \\
&=
\begin{pmatrix}
0 & \cdots & 0 & \beta_{1} & \cdots & \beta_{m} \\
\gamma_{1,1} & \cdots & \gamma_{1, m} & \alpha_{1,1} & \cdots & \alpha_{1, m} \\
\gamma_{2,1} & \cdots & \gamma_{2, m} & \alpha_{2,1} & \cdots & \alpha_{2, m} \\
\vdots & \ddots & \vdots & \vdots & \ddots & \vdots \\
\gamma_{m,1} & \cdots & \gamma_{m, m} & \alpha_{m,1} & \cdots & \alpha_{m, m}
\end{pmatrix}.
\end{align}

Consider the $k$-th network layer $\mathcal{T}^k$, it integrates self-attention layers or feed-forward networks within transformers. The output of the FC, denoted by $\mathbf{H}^{k}$, can be simply formulated as follows:
\begin{align}
\mathbf{H}^{k} &= \mathcal{FC}^k(\mathcal{T}^{k}, \mathbf{H}^{k-1}) \notag \\
&= {\mathbf{B}^k}^\intercal \mathcal{T}^k\big({\mathbf{Y}^k}^{\intercal} \mathbf{H}^{k-1}\big) 
+ {\mathbf{A}^k}^\intercal \mathbf{H}^{k-1}. \label{eq:hc_recurrent_form}
\end{align}

\subsection{Dynamic and Static Frac-Connections}
Since Frac-Connections is the fractional variant of Hyper-Connections, Frac-Connections can be implemented in two forms likewise:
\begin{enumerate}
    \item \textbf{Static Frac-Connections}: The weights are learnable, but static during testing.
    \item \textbf{Dynamic Frac-Connections}: The weights are dynamically computed based on the input, allowing greater flexibility.
\end{enumerate}

The matrix representation of dynamic frac-connections (DFC) is defined as follows:

\begin{equation}
\mathcal{FC}(\mathbf{H}) = \begin{pmatrix}
\mathbf{0}_{1\times m} & \mathcal{B}(\mathbf{H}) \\
\mathcal{Y}(\mathbf{H}) & \mathcal{A}(\mathbf{H})
\end{pmatrix}
\end{equation}

Similarly, given a layer \(\mathcal{T}\) and input \(\mathbf{H}\), we obtain the output of the DFC as follows:
\begin{equation}
\mathbf{\hat{H}} = \mathcal{FC}(\mathbf H)(\mathcal{T}, \mathbf H).
\end{equation}

In practice, we follow that of DHC~\cite{zhu2024hyper}, combining the dynamic and static matrices to achieve DFC. The dynamic parameters are obtained through a linear transformation. To stabilize the training process, we introduce normalization before the linear transformation and apply the tanh activation function after it, scaling it by a small initial learnable factor. 
The following equations detail how these dynamic parameters are computed:

\begin{align}
\label{eq:norm}
\overline{\mathbf{H}} &= \texttt{norm}(\mathbf{H}) \\
\label{eq:B}
\mathcal{B}(\mathbf{H})     &= 
s_\beta \circ \texttt{tanh} (\overline{\mathbf{H}} \mathbf{W}_{\beta})^\intercal + \mathbf{B} \in \mathbb{R}^{1\times m}\\
\label{eq:Y}
\mathcal{Y}(\mathbf{H}) &= 
s_\alpha \circ \texttt{tanh} (\overline{\mathbf{H}} \mathbf{W}_{\gamma}) + \mathbf{Y} \in \mathbb{R}^{m\times m} \\
\label{eq:A}
\mathcal{A}(\mathbf{H})     &= 
s_\alpha \circ \texttt{tanh} (\overline{\mathbf{H}} \mathbf{W}_{\alpha}) + \mathbf{A} \in \mathbb{R}^{m\times m} 
\end{align}

\subsection{Initialization and Implementation}
In order to make the initialization of the frac-connections equivalent to the Pre-Norm residual connections, we adopt the following initialization strategy. The dynamic parameters $\mathbf{W}_{\beta}$, $\mathbf{W}_{\gamma}$, and $\mathbf{W}_{\alpha}$ in Eqs.~\ref{eq:B}, \ref{eq:Y}, and \ref{eq:A} are initialized to 0, while the static matrices are initialized as follows:
\begin{equation}
\label{eq:initialization}
\begin{pmatrix}
\mathbf{0}_{1\times1} & \mathbf{B} \\
\mathbf{Y} & \mathbf{A}
\end{pmatrix}
=\begin{pmatrix}
\mathbf{0}_{1 \times 1} & \mathbf{1}_{1 \times m}\\
\mathbf{e}_{m\times m} & \mathbf{e}_{m\times m}
\end{pmatrix}.
\end{equation}

The static components $\mathbf{B}$, $\mathbf{Y}$, and $\mathbf{A}$ in Eqs.~\ref{eq:shc},~\ref{eq:A},~\ref{eq:B},~\ref{eq:Y} do not utilize weight decay, whereas the dynamic component does.

Frac-Connections for transformer is illuminated in Algorithm~\ref{alg:frac_connections} and Pytorch-style pseudocode is shown in Algorithm~\ref{alg:torch_fc}, \ref{alg:torch_trans_with_fc}.

\begin{algorithm}[h]
\caption{Frac-Connections for Transformers}\label{alg:frac_connections}
\begin{algorithmic}[1]
\Require Initial hidden vector $\mathbf{h}^0 \in \mathbb{R}^d$
\Require Fraction rate $m$
\Ensure Final output $\mathbf{y}$

\State \textbf{Initialize:}
\State $\mathbf{H}^0 \gets \texttt{Reshape}\big(\mathbf{h}^0, (m, d/m)\big)^\intercal \in \mathbb{R}^{m \times (d/m)}$

\For{$k = 1$ to $L$}
    \State $\mathbf{h_0}^{k-1} \gets \texttt{Reshape}({\mathbf{Y}^k}^\intercal\mathbf{H}^{k-1}, (d,))$ 
    \State $\mathbf{H}^k \gets {\mathbf{B}^k}^\intercal \texttt{Reshape}\big(\mathcal{T}^k(\mathbf{h_0}^{k-1}),(m, d/m)\big) + {\mathbf{A}^k}^\intercal \mathbf{H}^{k-1}$
\EndFor

\State $\mathbf{h}^L \gets \texttt{Reshape}\big(\mathbf{H}^L, (m, d/m)\big)$
\State $\mathbf{y} \gets \texttt{Umembedding}\big(\texttt{Norm}(\mathbf{h}^L)\big)$

\State \Return $\mathbf{y}$
\end{algorithmic}
\end{algorithm}

\subsection{Parameters and Computation}
\label{app:para_comp_mem_analysis}
{\bf Static Frac-Connections.} All learnable parameters are included in the frac-connection matrix $\mathcal{FC}$ in Eq.~\ref{eq:shc}. The number of parameters in one $\mathcal{FC}$ is given by:
\begin{equation}
    \left|\theta_{\texttt{SHC}}\right|= |\theta_{\mathbf{B}}| + |\theta_{\mathbf{Y}}| + |\theta_{\mathbf{A}}|= m + m \cdot m + m \cdot m=m \cdot (2m+1).
\end{equation}
Thus, the number of extra parameters is:
\begin{equation}
P_{\texttt{extra}}=\left|\theta_{\texttt{SHC}}\right| \times 2 \times L,
\end{equation}
where $L$ is the number of layers. For example, in \texttt{OLMo-1B-7B-SFC$\times$4}, $P_{\texttt{extra}}=1152$.

{\bf Dynamic Frac-Connections.} The parameters of DHC are defined in Eqs.~\ref{eq:norm}, \ref{eq:B}, \ref{eq:Y}, and \ref{eq:A}, and the number of parameters is given by:
\begin{align}
    \left|\theta_{\texttt{DFC}}\right| &= |\theta_{\texttt{norm}}| + |s_{\beta}| + |\theta_{\mathbf{W}_\beta}| + |\theta_{\mathbf{B}}| + 
    |s_{\alpha}| + |\theta_{\mathbf{W}_{\gamma}}| + |\theta_{\mathbf{Y}}| +
    |\theta_{\mathbf{W}_{\alpha}}| + |\theta_{\mathbf{A}}| \\
    &=|\theta_{\texttt{norm}}| + d_{\texttt{model}}/m\times(2m+1) + m \cdot (2m+1) + 2,
\end{align}
where $d_{\texttt{model}}$ is the dimension of the hidden states in the transformer, and $|\theta_{\texttt{norm}}|$ depends on the type of normalization module. For RMSNorm~\cite{zhang2019root}, $|\theta_{\texttt{norm}}|=d_{\texttt{model}}/m$. Similar to the static hyper-connections, the number of extra parameters is:
\begin{equation}
P_{\texttt{extra}}=\left|\theta_{\texttt{DFC}}\right| \times 2 \times L,
\end{equation}
For example, for \texttt{OLMo-1B-7B-DFC$\times$4}, $P_{\texttt{extra}}==165,056$. The number of parameters for \texttt{DFC} used in the experiments is detailed in Table~\ref{tab:params_comparison}.
\begin{table}[h]
\centering
\caption{Comparison of number of parameters.}
\begin{tabular}{lccc}
\toprule
\textbf{Method}       & \makecell{\textbf{FC Params(B)}} & \makecell{\textbf{Total Params(B)}} & \makecell{\textbf{Total Params $\Delta$ rate (\%)}} \\ 
\midrule
OLMo-1B2             & -                   & 1.17676442  & - \\ 
OLMo-1B2-DFC$\times$4       & 0.000165     & 1.17715846  & \bf{+0.014\%} \\ 
\midrule  % Add a thicker line between 1B and 7B
OLMoE-1B-7B         & -                    & 6.91909427  & - \\ 
OLMoE-1B-7B-DFC$\times$4   & 0.000165            & 6.91948832  & \bf{+0.0024\%} \\ 
\bottomrule
\end{tabular}
\label{tab:params_comparison}
\end{table}

\textbf{Computational Analysis.} 
The primary computational cost of both SFC and DFC occurs in line 5 of Algorithm~\ref{alg:frac_connections}, with a complexity of $\mathcal{O}(d_{\text{model}} \times 4m)$. For comparison, the computational cost of the Feed-Forward Network (FFN) is $\mathcal{O}(2 \times d_{\text{model}} \times d_{\text{ffn}})$, while the projection component of attention requires $\mathcal{O}(4 \times d_{\text{model}} \times d_{\text{model}})$ operations.

Since $\mathcal{O}(d_{\text{model}} \times 4m) \ll \mathcal{O}(4 \times d_{\text{model}} \times d_{\text{model}}) < \mathcal{O}(2 \times d_{\text{model}} \times d_{\text{ffn}})$, the computational overhead of FC implementations is negligible compared to the costs of both the FFN and attention projection operations. Here, $d_{\text{ffn}}$ represents the inner dimension of the FFN. 
Our analysis confirms that regardless of whether \texttt{SFC} or \texttt{DFC} is implemented, both the additional parameters and computational overhead introduced remain minimal and can be considered negligible in the overall system performance. Detailed computational cost statistics of \texttt{DFC} are presented in Table~\ref{tab:flops_comparison}. 
\begin{table}[h]
\centering
\caption{FLOPs per token in forward pass.}
\begin{tabular}{lccc}
\toprule
\textbf{Method}       & \makecell{\textbf{FC FLOPs (G)}} & \makecell{\textbf{Total FLOPs (G)}} & \makecell{\textbf{Total FLOPs $\Delta$ rate (\%)}} \\ 
\midrule
OLMo-1B               & -              & 2.5587  & - \\ 
OLMo-1B-DFC$\times$4         & 0.0013         & 2.5598  & \textbf{+0.044\%} \\ 
\midrule
OLMoE-1B-7B           & -              & 2.3580  & - \\ 
OLMoE-1B-7B-DFC$\times$4     & 0.0013   & 2.3629  & \textbf{+0.056\%} \\ 
\bottomrule
\end{tabular}
\label{tab:flops_comparison}
\end{table}

\section{Experiments}
We evaluate Frac-Connections on the pre-training of large language models, including sparse and dense models. Specifically, for sparse models we study Sparse Mixture-of-Experts (MoE) models~\cite{shazeer2017sparsely} and follow the experimental setup described by OLMoE~\cite{muennighoff2024olmoeopenmixtureofexpertslanguage}, conducting ablation studies on \texttt{OLMoE-1.3B}, which has 1.3B total parameters with 260M activated parameters. We further validate the effectiveness of our approach on a larger sparse model, \texttt{OLMoE-7B}, which has 7B total parameters with 1.3B activated parameters. For dense models, we follow the OLMo2~\cite{olmo20242olmo2furious} training setup to pre-train a 1B2 parameter model. Importantly, all experiments were conducted without hyperparameter tuning, and the training hyperparameters were strictly aligned across comparative baselines. Through these experiments across different model scales and architectures, we aim to comprehensively demonstrate the applicability and benefits of our proposed Frac-Connections approach.

\subsection{Ablation Study} 

\begin{figure}[h]
    \centering
    \begin{minipage}{0.33\textwidth}
        \centering
        \includegraphics[width=\linewidth]{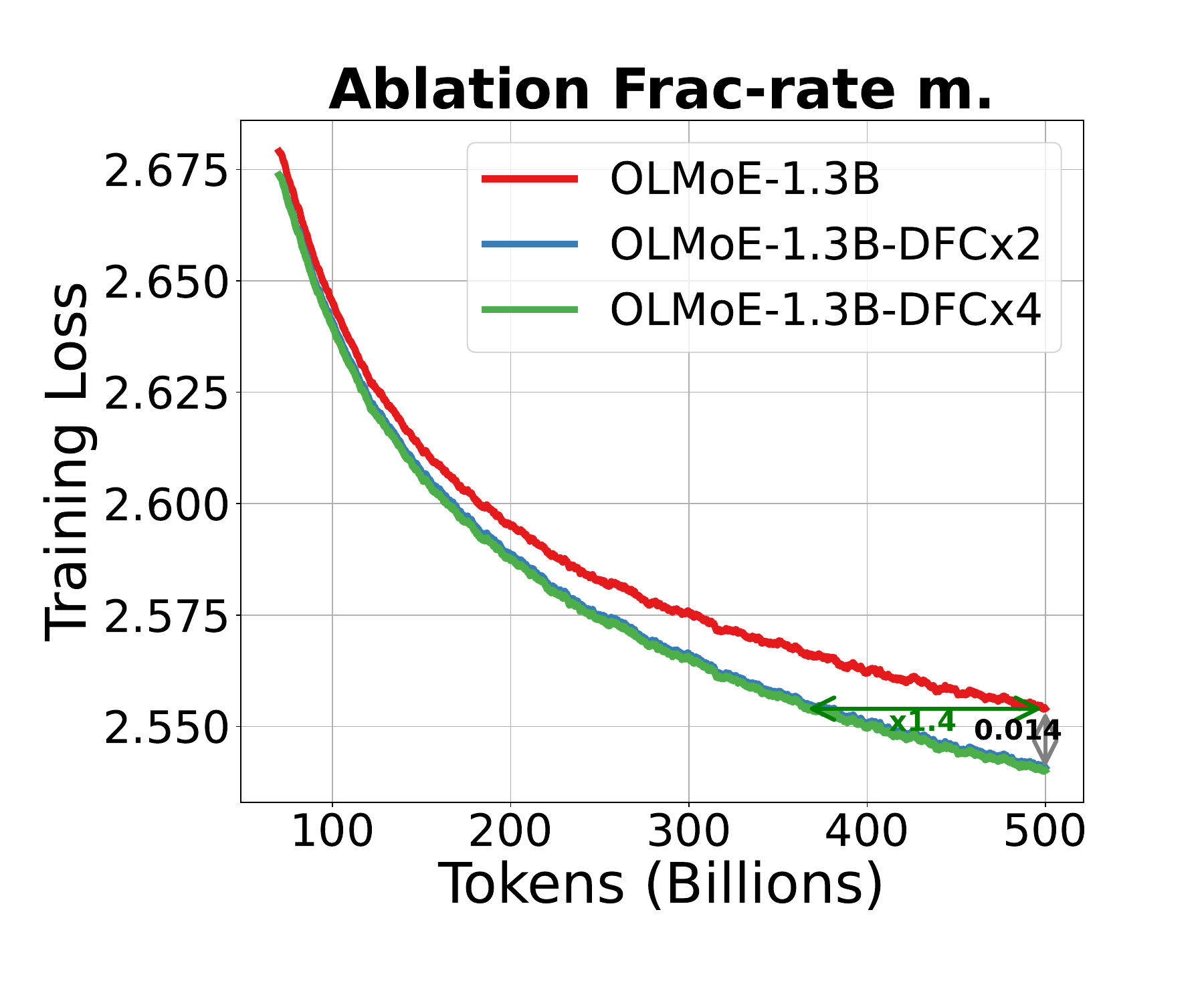}
        
    \end{minipage}\hfill
    \begin{minipage}{0.33\textwidth}
        \centering
        \includegraphics[width=\linewidth]{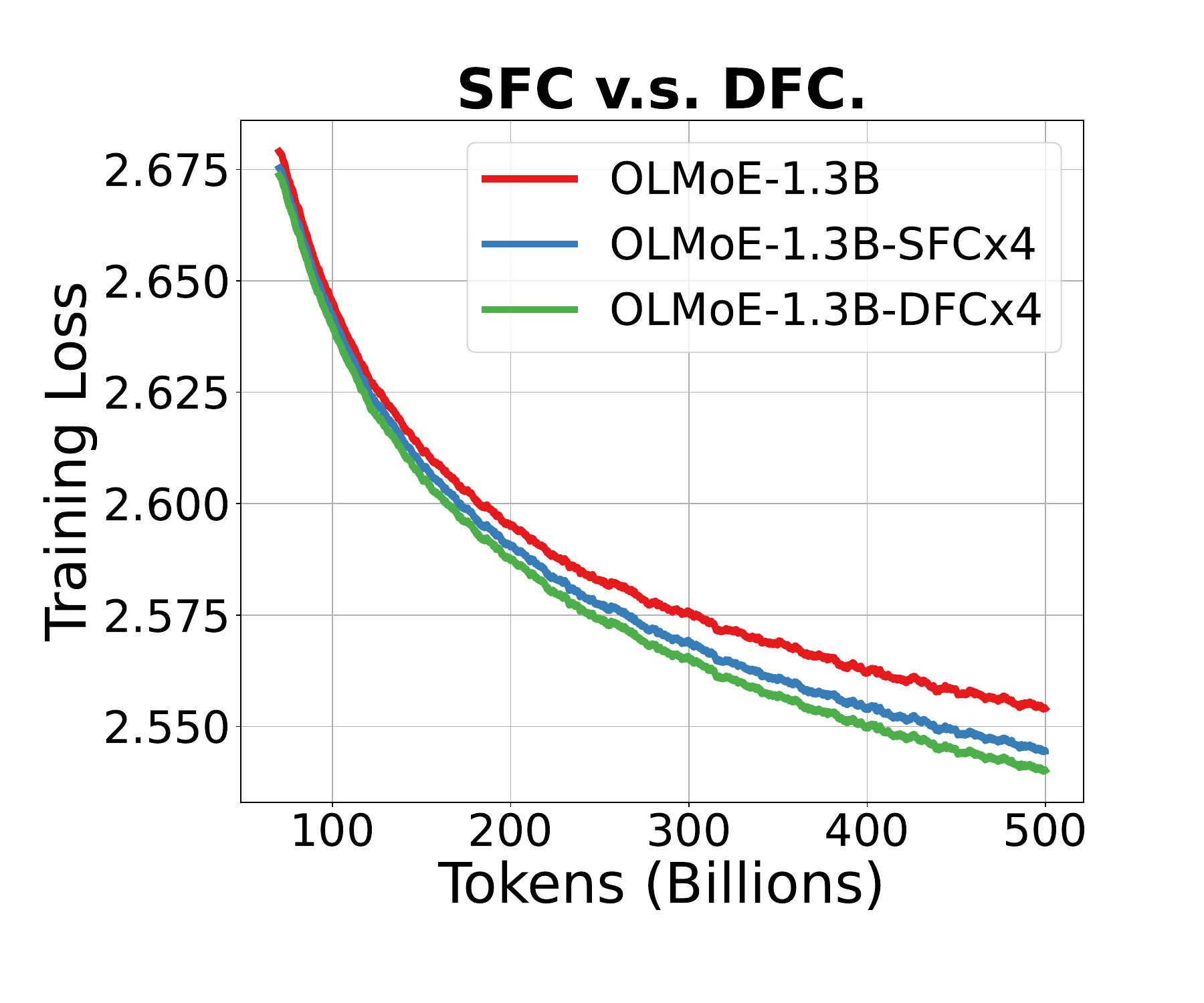}

     \end{minipage}\hfill
    \begin{minipage}{0.33\textwidth}
        \centering
        \includegraphics[width=\linewidth]{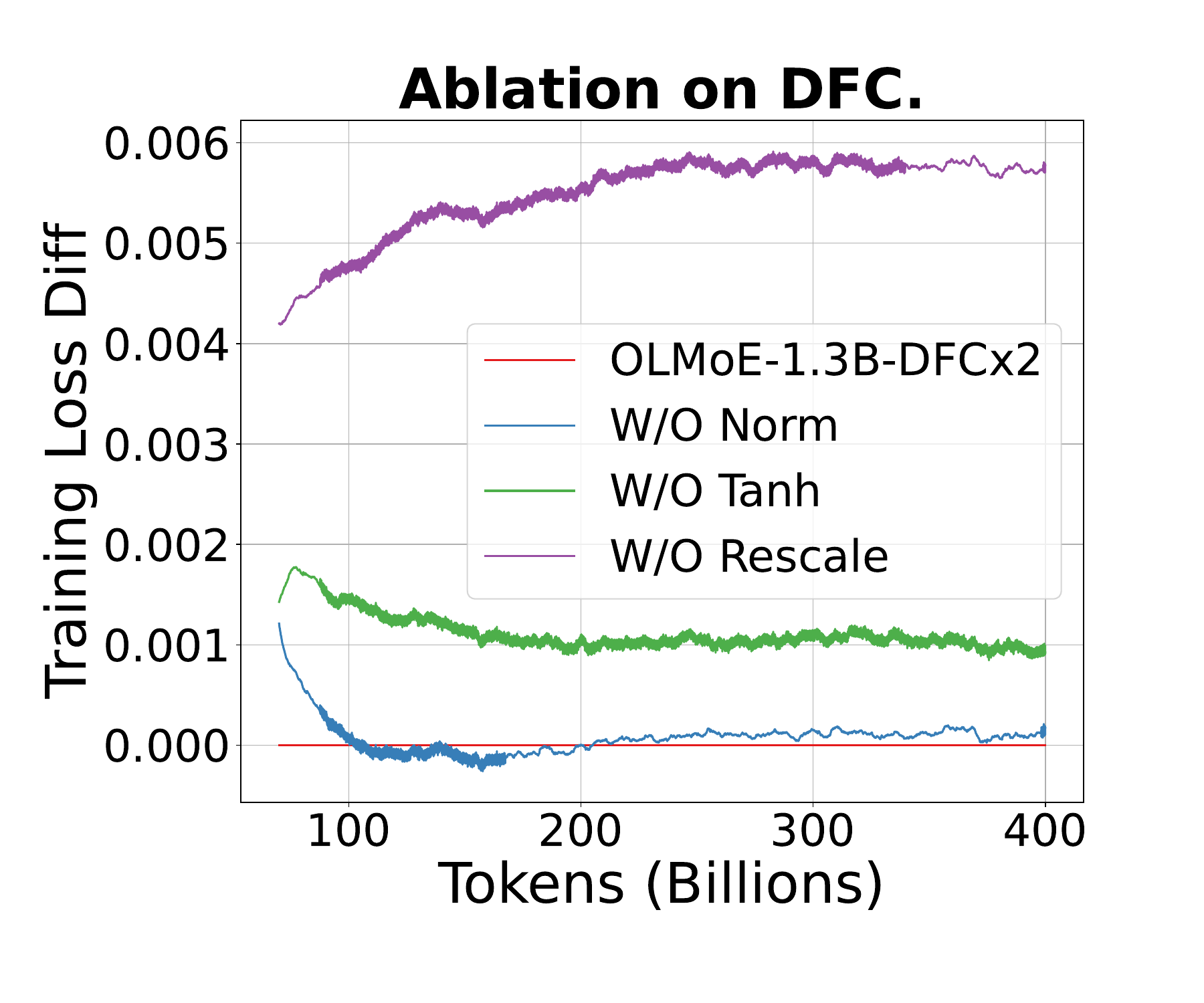}

     \end{minipage}
    \caption{Training loss (0.999 EMA smoothed) loss for \texttt{OLMoE-1.3B} models.}
    \label{fig:olmoe1b3_training_loss}
\end{figure}

We conduct extensive ablation studies on the \texttt{OLMoE-1.3B} model to evaluate different configurations of Frac-Connections, as shown in Figure~\ref{fig:olmoe1b3_training_loss}.

{\bf Effect of different frac-rates.} The leftmost of Figure~\ref{fig:olmoe1b3_training_loss} compares the baseline model against versions with Dynamic Frac-Connections (DFC) at different frac-rates (DFC$\times$2 and DFC$\times$4). The results show that DFC$\times$2 demonstrates significant improvement over the baseline, while DFC$\times$4 offers only marginal additional gains compared to DFC$\times$2. The \texttt{OLMoE-1.3B-DFC$\times$4} model exhibits a training loss reduction of approximately {\bf0.014} compared to the baseline.

{\bf Static Frac-Connections (SFC) v.s. Dynamic Frac-Connections (DFC) }. In the middle of Figure~\ref{fig:olmoe1b3_training_loss}, both \texttt{-SFC$\times$4} and \texttt{-DFC$\times$4} outperform the baseline. Additionally, \texttt{-DFC$\times$4} achieves better results than \texttt{-SFC$\times$4}, suggesting that the dynamic parameter prediction mechanism provides additional modeling capacity.

{\bf Ablation study on the components of DFC.} The rightmost of Figure~\ref{fig:olmoe1b3_training_loss} evaluates the impact of normalization, tanh activation, and rescaling by measuring their loss differences relative to the \texttt{-DFC×2} baseline. 
From the training loss perspective, removing rescaling (purple line, without $s_{\beta}$ and $s_{\alpha}$ in Eq.~\ref{eq:B},~\ref{eq:Y},~\ref{eq:A}) causes the most severe performance degradation, followed by the removal of tanh activation (green line), while the absence of normalization (blue line) results in the least detrimental effect, though still negatively impacting performance. These findings demonstrate the hierarchical importance of each component in the DFC implementation, with rescaling being particularly crucial for maintaining optimal training dynamics. Given that the original DHC design components exhibit either substantial or modest improvements when implemented in DFC, we opt to preserve the complete original DHC architecture to maintain optimal performance characteristics.

These findings underscore the effectiveness of Frac-Connections as a lightweight yet impactful enhancement for transformer-based models, offering improved performance with minimal parameter overhead.

\subsection{MoE Models}

\begin{figure}[h]
    \centering
    \begin{minipage}{0.25\textwidth}
        \centering
        \includegraphics[width=\linewidth]{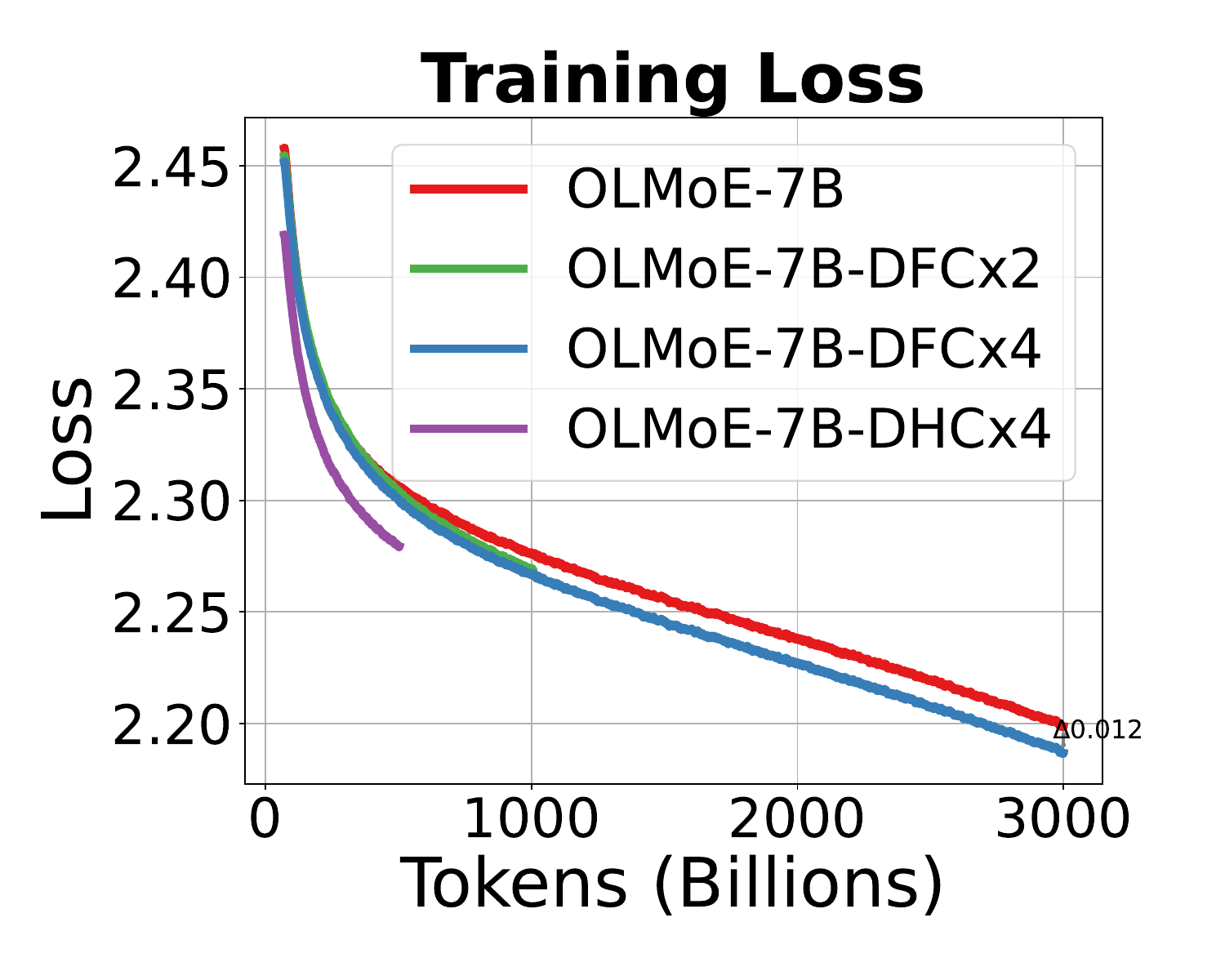}
        
    \end{minipage}\hfill
    \begin{minipage}{0.25\textwidth}
        \centering
        \includegraphics[width=\linewidth]{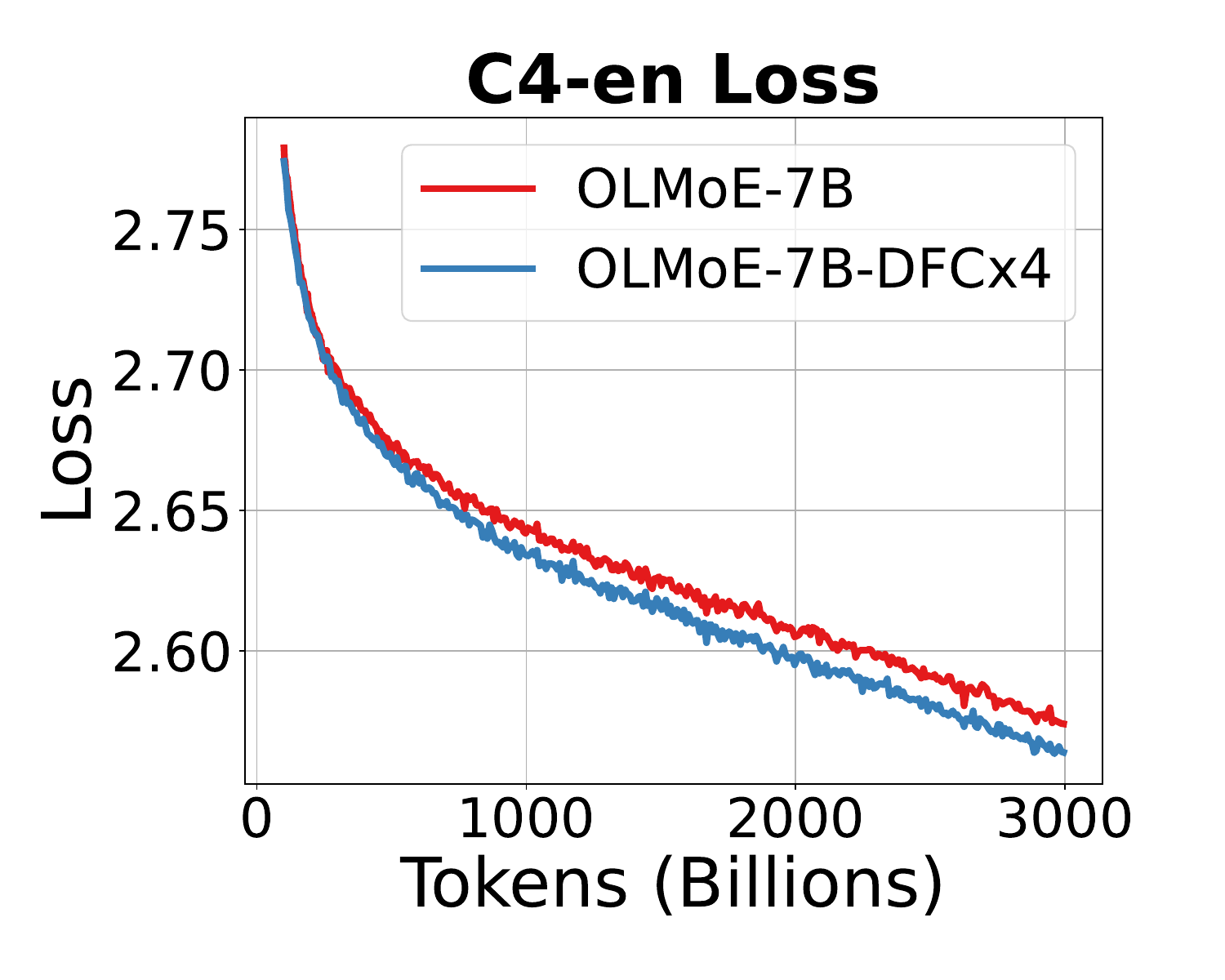}

     \end{minipage}\hfill
    \begin{minipage}{0.25\textwidth}
        \centering
        \includegraphics[width=\linewidth]{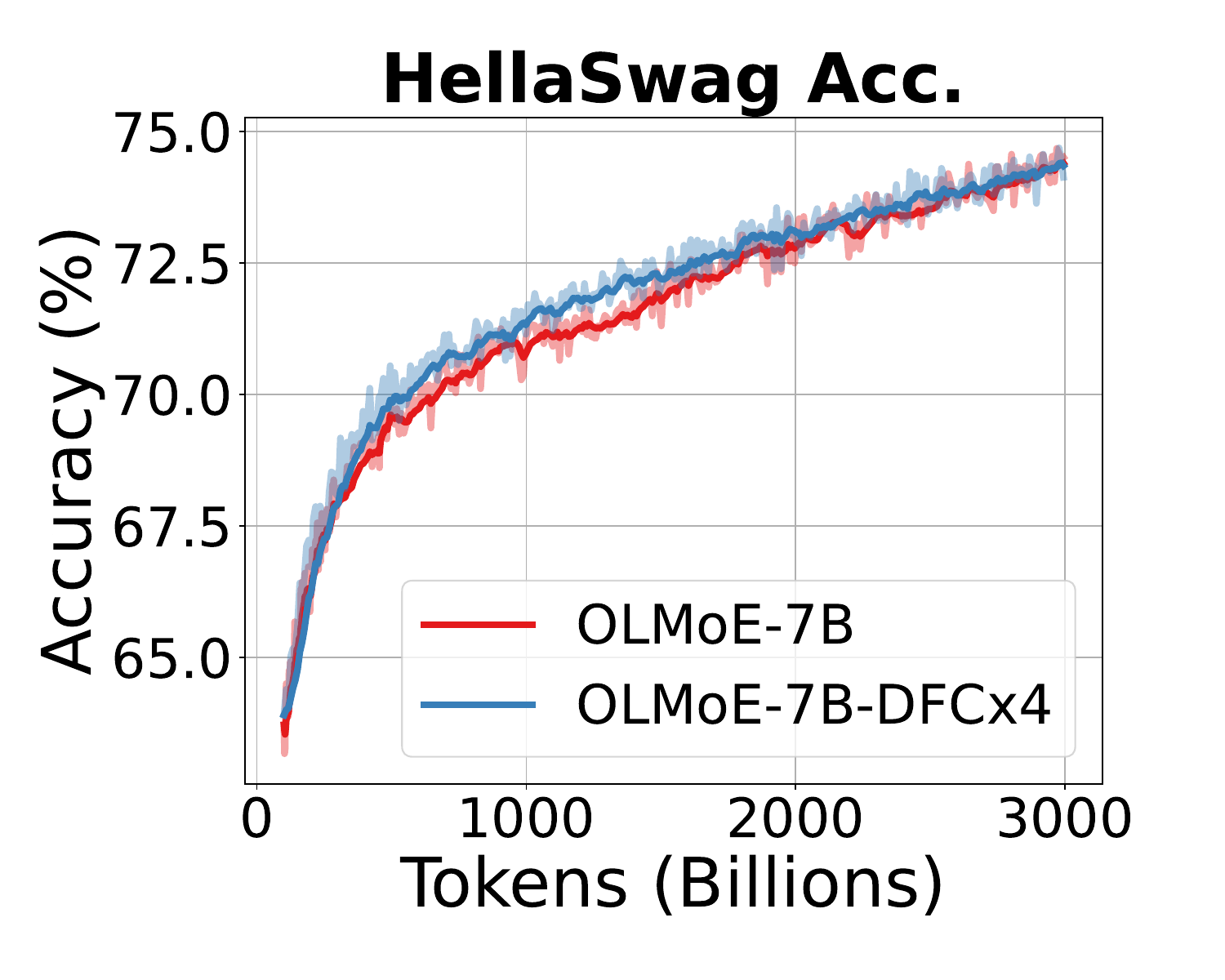}

     \end{minipage}\hfill
    \begin{minipage}{0.25\textwidth}
        \centering
        \includegraphics[width=\linewidth]{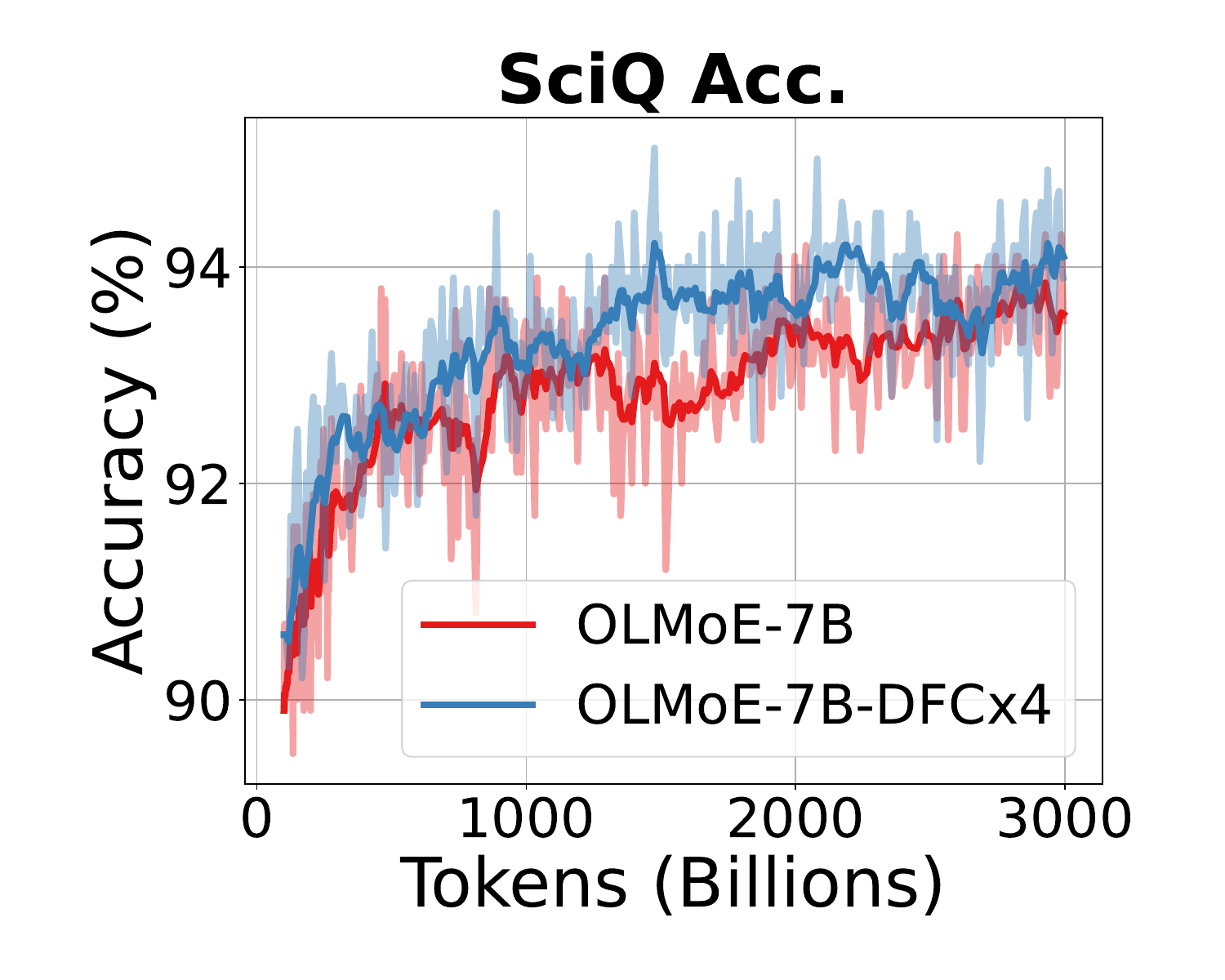}

     \end{minipage} \\
    \begin{minipage}{0.25\textwidth}
        \centering
        \includegraphics[width=\linewidth]{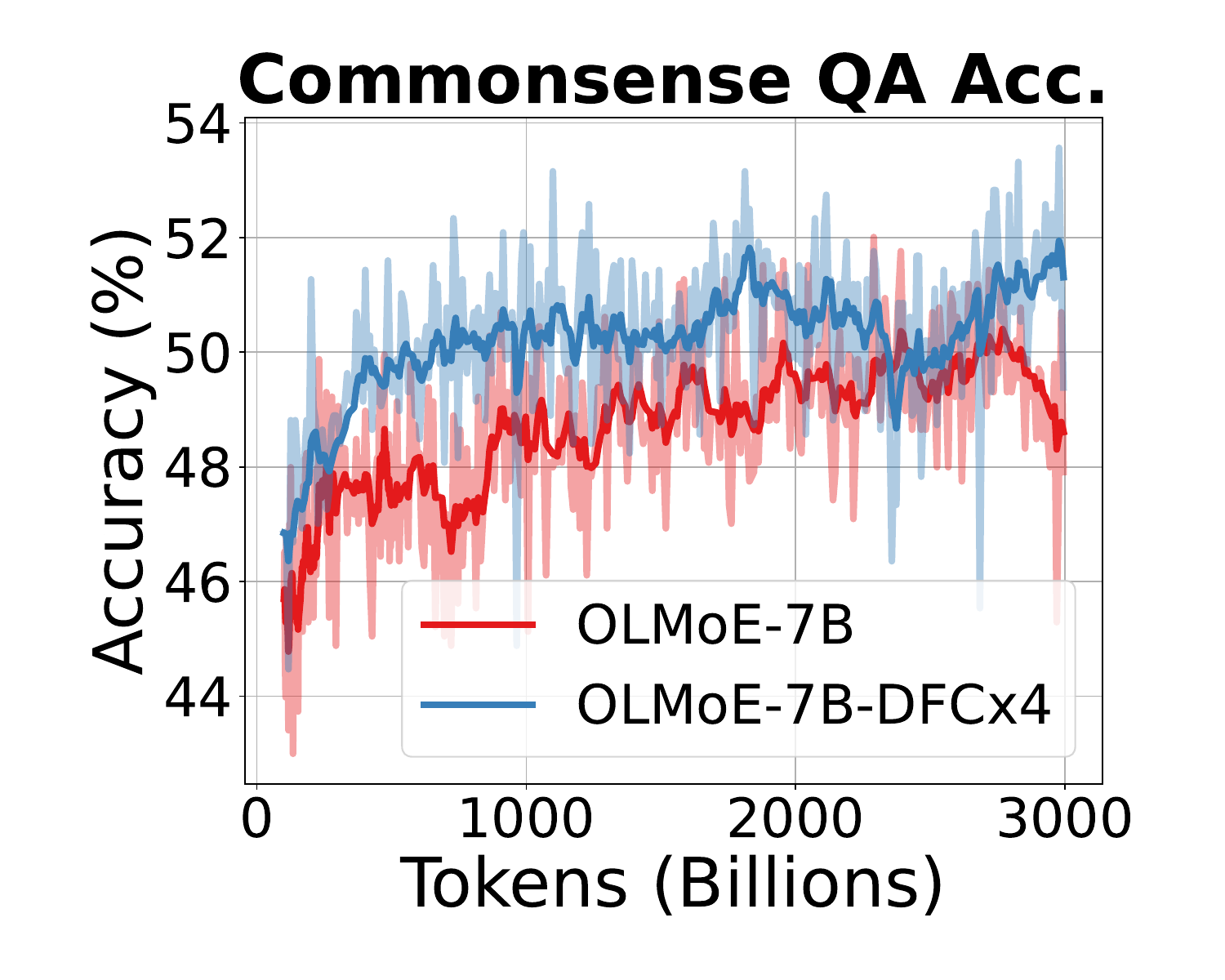}
        
    \end{minipage}\hfill
    \begin{minipage}{0.25\textwidth}
        \centering
        \includegraphics[width=\linewidth]{fig/olmoe_7b_SciQ.pdf}

     \end{minipage}\hfill
    \begin{minipage}{0.25\textwidth}
        \centering
        \includegraphics[width=\linewidth]{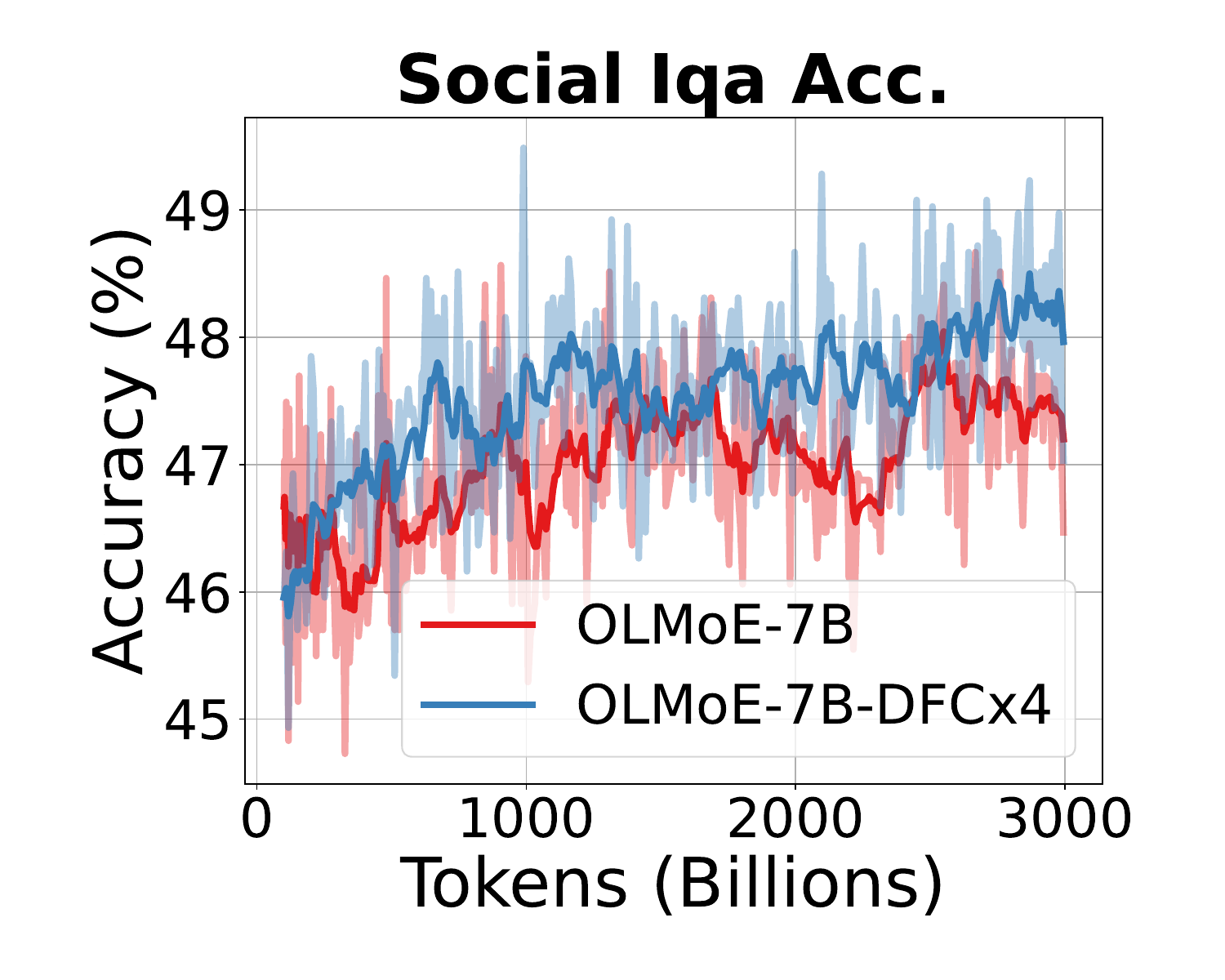}

     \end{minipage}\hfill
    \begin{minipage}{0.25\textwidth}
        \centering
        \includegraphics[width=\linewidth]{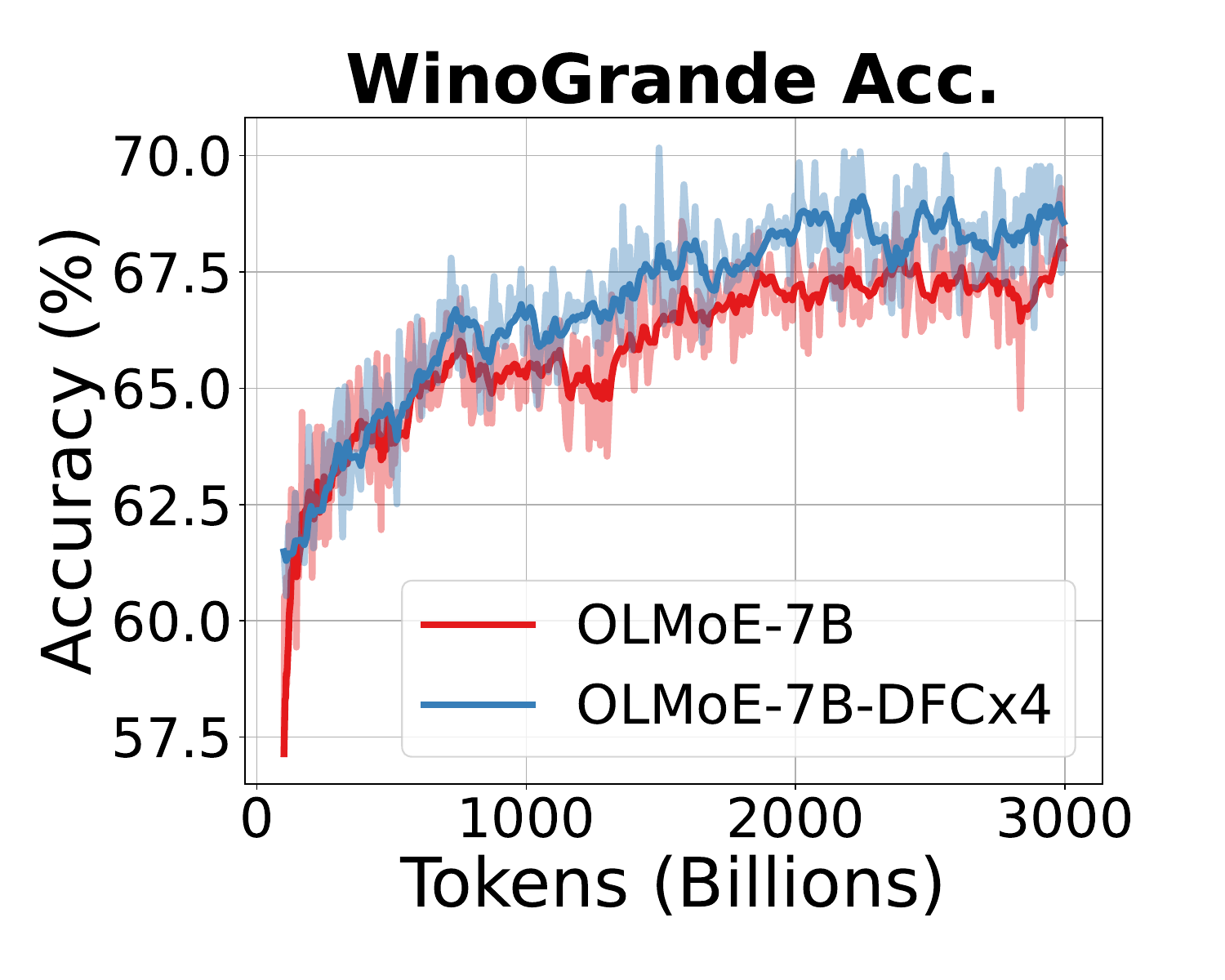}

     \end{minipage}
    \caption{Training and evaluation performance of \texttt{OLMoE-7B} models. The plots show the training loss, C4-en loss, and accuracy on HellaSwag, SciQ, Commonsense QA, Social IQA, and WinoGrande over the course of training. The results are EMA-smoothed for clarity. The \texttt{OLMoE-7B-DFCx4} variant demonstrates improved loss reduction and higher accuracy across multiple benchmarks compared to the baseline \texttt{OLMoE-7B} model, indicating enhanced optimization efficiency and generalization.}
    \label{fig:olmoe_training_loss}
\end{figure}

{\bf Converge curves.} As shown in Figure~\ref{fig:olmoe_training_loss}, from the training loss and C4-en loss curves, we observe that \texttt{OLMoE-7B-DFC$\times$4} achieves a faster convergence, with a reduction of \textbf{0.012} in training loss compared to the baseline. Furthermore, we observe that Hyper-Connections (\texttt{OLMoE-7B-DHC$\times$4}) converge significantly faster than Frac-Connections (\texttt{OLMoE-7B-DFC$\times$4}), suggesting that when applying HC or FC, a trade-off between memory consumption and performance needs to be considered. 

\begin{table}[h!]
\centering
\caption{Downstream evaluations for OLMoE-7B models with training 3T tokens. MMLU Var is a modified version of MMLU that includes varying few-shot examples, providing stable feedback during early training.}
\resizebox{\textwidth}{!}{
\begin{tabular}{lcccccccc}
\toprule
\textbf{Method} & 
\textbf{\begin{tabular}[c]{@{}c@{}}Hella-\\Swag\end{tabular}} & 
\textbf{BoolQ} & 
\textbf{\begin{tabular}[c]{@{}c@{}}Wino-\\Grande\end{tabular}} &  
\textbf{\begin{tabular}[c]{@{}c@{}}MMLU\\ Var\end{tabular}} &  
\textbf{PIQA} & 
\textbf{SciQ} &  
\textbf{\begin{tabular}[c]{@{}c@{}}Common-\\sense QA\end{tabular}} &  
\textbf{AVG} \\ 
\midrule
OLMoE-7B & 74.28 & \textbf{72.87} & 67.64 & 41.83 & 78.73 & 93.60 & 49.14 & 68.30 \\ 
OLMoE-7B-DFC$\times$4 & \textbf{74.48} & 72.11 & \textbf{68.59} & \textbf{42.33} & \textbf{79.16} & \textbf{94.10} & \textbf{49.80} & \textbf{68.65} \\ 
\bottomrule
\end{tabular}
}
\label{tab:olmoe_results}
\end{table}

{\bf Downstream performance.} Throughout training, the \texttt{OLMoE-7B-DFC$\times$4} variant maintains a consistent advantage on most benchmarks, including Commonsense QA and WinoGrande QA. For HellaSwag, the \texttt{OLMoE-7B-DFC$\times$4} variant maintains an early advantage over the baseline; however, as training progresses, the gap narrows, and the baseline model nearly catches up toward the end. Table~\ref{tab:olmoe_results} shows the performance of models trained with 3T tokens, and the \texttt{OLMoE-7B-DFC$\times$4} variant demonstrates higher accuracy across most benchmarks. Specifically, it outperforms the baseline by \textbf{+0.95\%} on WinoGrande (67.64\% $\rightarrow$ 68.59\%), \textbf{+0.50\%} on MMLU Var (41.83\% $\rightarrow$ 42.33\%), and \textbf{+0.66\%} on Commonsense QA (49.14\% $\rightarrow$ 49.80\%), indicating that Frac-Connections enhance knowledge retention and generalization. 

These results indicate that Frac-Connections not only improve training efficiency but also lead to better model generalization across diverse NLP tasks.

\subsection{Dense Models}

\begin{figure}[h]
    \centering
    \begin{minipage}{0.25\textwidth}
        \centering
        \includegraphics[width=\linewidth]{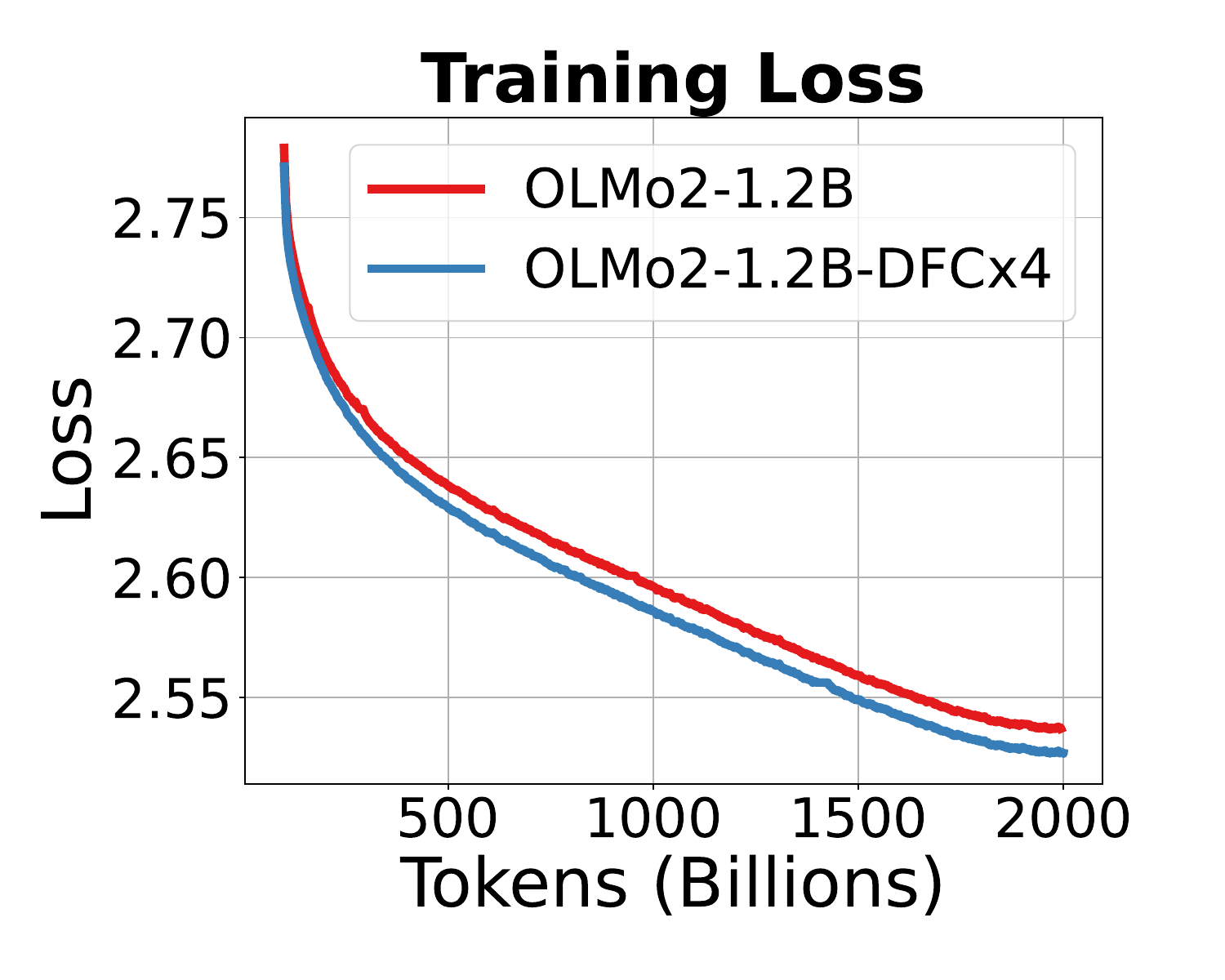}
        
    \end{minipage}\hfill
    \begin{minipage}{0.25\textwidth}
        \centering
        \includegraphics[width=\linewidth]{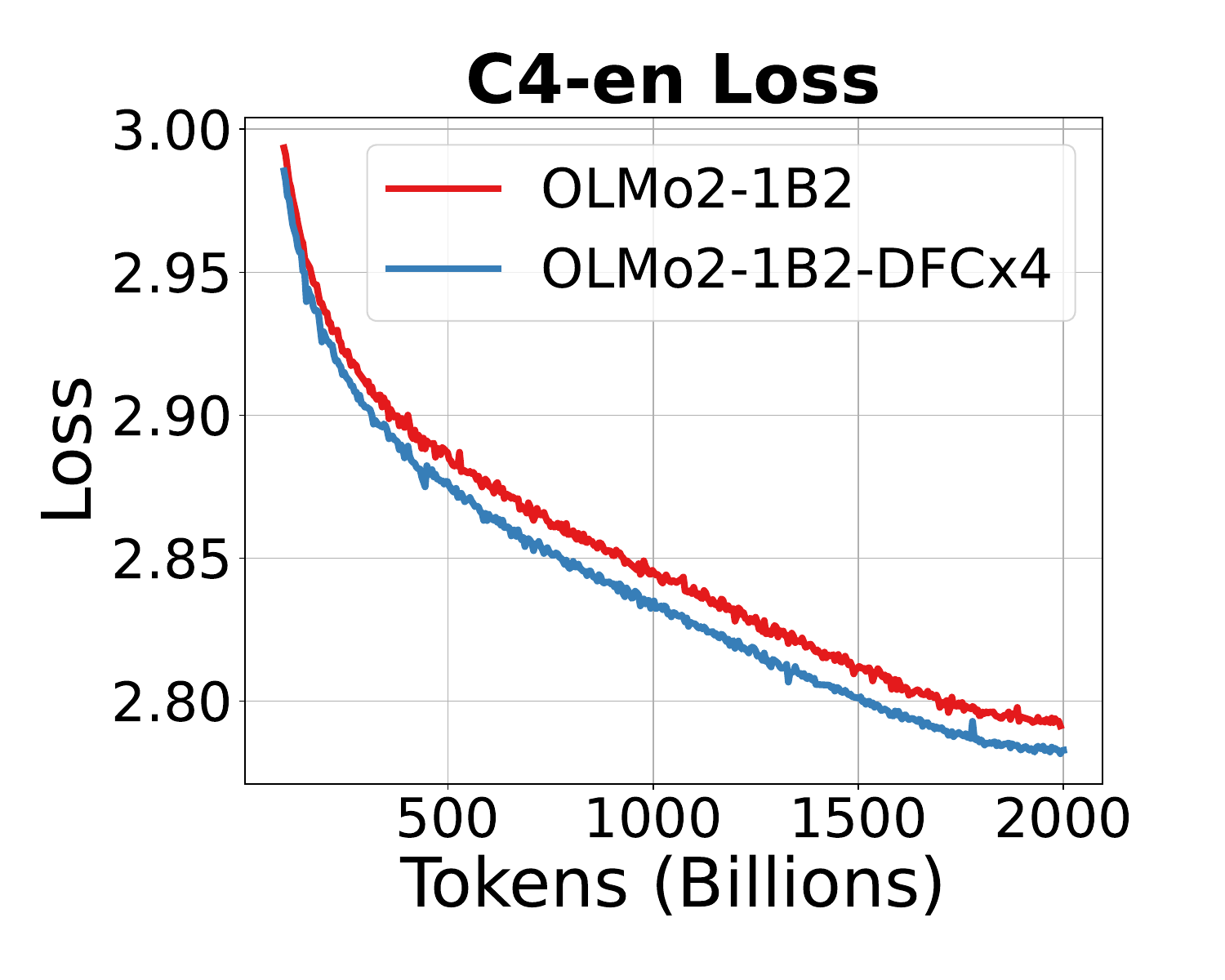}

     \end{minipage}\hfill
    \begin{minipage}{0.25\textwidth}
        \centering
        \includegraphics[width=\linewidth]{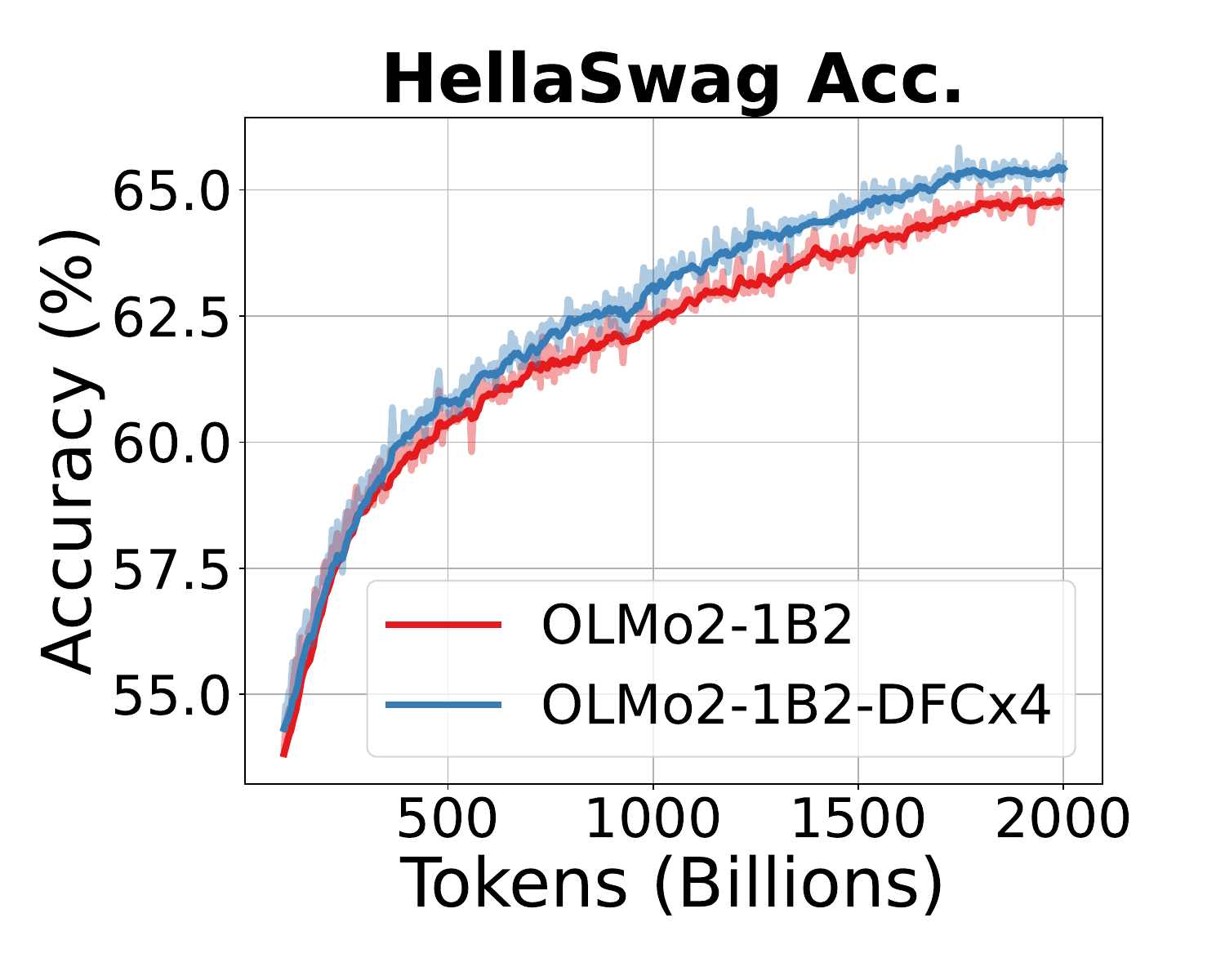}

     \end{minipage}\hfill
    \begin{minipage}{0.25\textwidth}
        \centering
        \includegraphics[width=\linewidth]{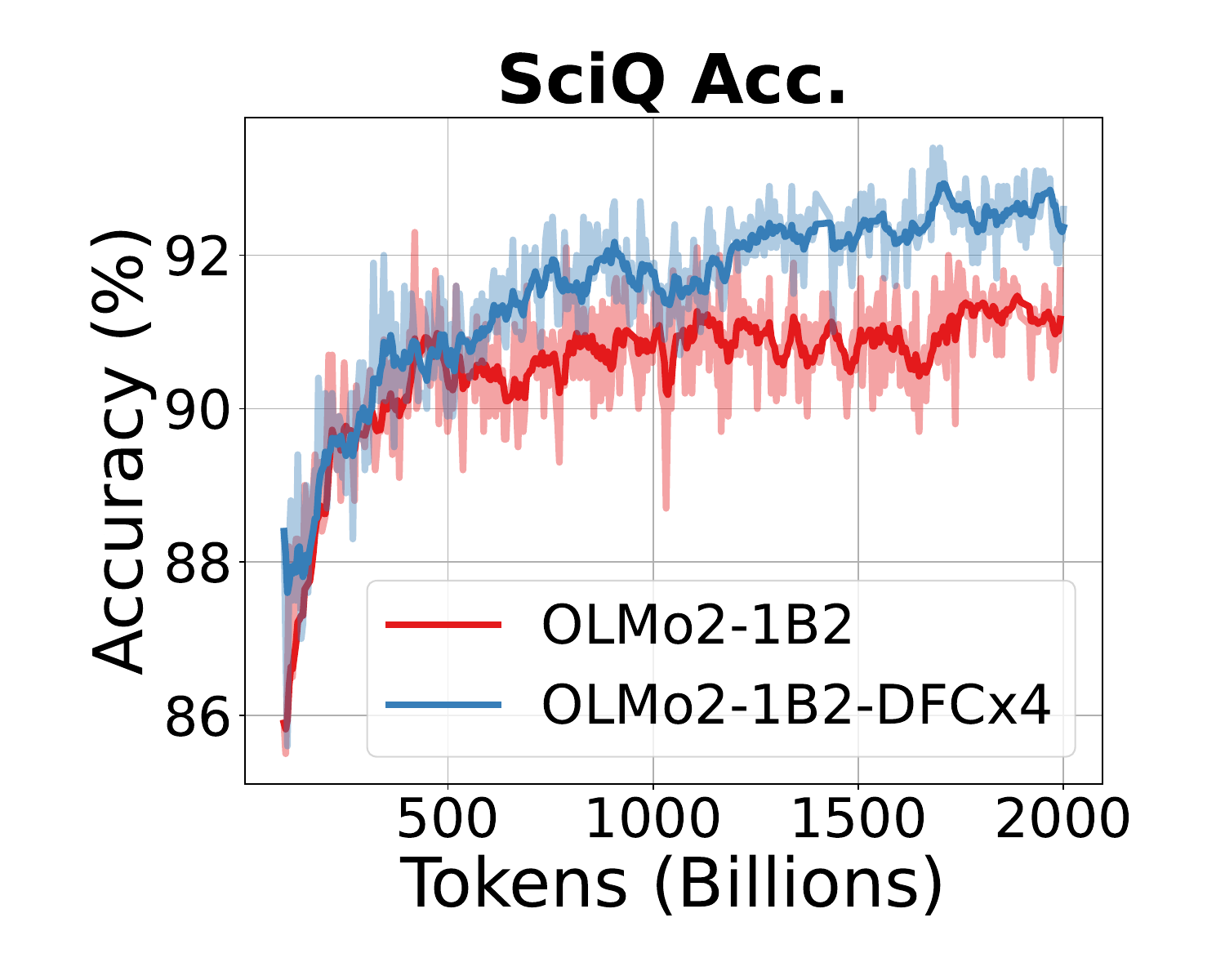}

     \end{minipage}
    \caption{Training and evaluation performance of \texttt{OLMo2-1B2} models. The plots show the training loss, C4-en loss, and accuracy on HellaSwag and SciQ over the course of training. The results are EMA-smoothed for clarity. The \texttt{OLMo2-1B2-DFCx4} variant demonstrates improved loss reduction and higher accuracy compared to the baseline OLMo2-1B2 model.}
    \label{fig:olmo2_training_loss}
\end{figure}

We evaluate Frac-Connections through experiments on the \texttt{OLMo2-1B2} model, as illustrated in Figure~\ref{fig:olmo2_training_loss} and Table~\ref{tab:olmo2_results}. \texttt{OLMo2-1B2-DFC$\times$4} variant exhibits consistently lower training loss and C4-en loss compared to the baseline \texttt{OLMo2-1B2} model. Furthermore, the \texttt{OLMo2-1B2-DFC$\times$4} variant consistently outperforms the baseline on HellaSwag and SciQ throughout training. This suggests that Frac-Connections facilitate more efficient optimization and improving parameter utilization.

\begin{table}[h!]
\centering
\caption{Downstream evaluations for OLMo2 models with training 2T tokens. MMLU Var is a modified version of MMLU that includes varying few-shot examples, providing stable feedback during early training.}
\resizebox{\textwidth}{!}{
\begin{tabular}{lcccccccc}
\toprule
\textbf{Methods} & 
\textbf{\begin{tabular}[c]{@{}c@{}}Hella-\\Swag\end{tabular}} & 
\textbf{BoolQ} & 
\textbf{\begin{tabular}[c]{@{}c@{}}Wino-\\Grande\end{tabular}} &  
\textbf{\begin{tabular}[c]{@{}c@{}}MMLU\\ Var\end{tabular}} &  
\textbf{PIQA} & 
\textbf{SciQ} &  
\textbf{\begin{tabular}[c]{@{}c@{}}Common-\\sense QA\end{tabular}} &  
\textbf{AVG} \\ 
\midrule
OLMo2-1B2 & 64.7 & 63.0 & 61.6 & 36.4 & \textbf{75.6} & 91.8 & 44.6 & 62.5 \\ 
OLMo2-1B2-DFC$\times$4 & \textbf{65.4} & \textbf{65.1} & \textbf{62.7} & \textbf{37.1} & 75.2 & \textbf{92.2} & \textbf{44.7} & \textbf{63.2} \\ 
\bottomrule
\end{tabular}
}
\label{tab:olmo2_results}
\end{table}

The downstream evaluation results in Table~\ref{tab:olmo2_results} demonstrate that \texttt{OLMo2-1B2-DFC$\times$4} achieves superior performance across multiple tasks, particularly on BoolQ (+2.1\%), WinoGrande (+1.1\%), and SciQ (+0.4\%), while maintaining comparable performance on PIQA. The average accuracy improvement of +0.7\% confirms that Frac-Connections enhance generalization across diverse benchmarks. Notably, the improvements on reasoning-intensive tasks such as BoolQ and WinoGrande highlight the ability of Frac-Connections to enhance model expressiveness without increasing computational overhead.

\section{Conclusion}
We introduced Frac-Connections, an efficient alternative to Hyper-Connections that divides the hidden states into fractions rather than expanding their width. Frac-Connections address the seesaw effect between gradient vanishing and representation collapse while reducing memory usage and computational costs. Our experimental results demonstrate that Frac-Connections are a practical and scalable solution for large language models.

% \clearpage

\bibliographystyle{plainnat}
\bibliography{main}

\clearpage

\beginappendix

\section{PyTorch Implementation of Frac-connections}

\begin{algorithm}[H]
\caption{Pseudocode of frac-connections in a PyTorch-like style.}
\label{alg:torch_fc}
\algcomment{\fontsize{7.2pt}{0em}\selectfont 
%\vspace{-1.em}
}
\definecolor{codeblue}{rgb}{0.25,0.5,0.5}
\lstset{
  backgroundcolor=\color{white},
  basicstyle=\fontsize{7.2pt}{7.2pt}\ttfamily\selectfont,
  columns=fullflexible,
  breaklines=true,
  captionpos=b,
  commentstyle=\fontsize{7.2pt}{7.2pt}\color{codeblue},
  keywordstyle=\fontsize{7.2pt}{7.2pt},
%  frame=tb,
}
\begin{lstlisting}[language=python]
# h: hidden vector (BxLxD)

class FracConnection(nn.Module):
    def __init__(self, dim, rate, config, dynamic_alpha, dynamic_beta, device=None):
        super(FracConnection, self).__init__()

        self.rate = rate
        self.dynamic_alpha = dynamic_alpha
        self.dynamic_beta = dynamic_beta
        self.use_tanh = config.use_tanh
        self.use_hc_norm = config.use_hc_norm
        self.use_scale = config.use_scale

        self.static_beta = nn.Parameter(torch.ones((rate,), device=device))
        self.static_alpha = nn.Parameter(torch.cat([torch.eye((rate), device=device), torch.eye((rate), device=device)], dim=1))

        if self.dynamic_alpha:
            self.dynamic_alpha_fn = nn.Parameter(torch.zeros((dim // self.rate, rate*2), device=device))

        if self.dynamic_beta:
            self.dynamic_beta_fn = nn.Parameter(torch.zeros((dim // self.rate, ), device=device))
        
        if self.use_scale:
            self.dynamic_alpha_scale = nn.Parameter(torch.ones(1, device=device) * 0.01)
            self.dynamic_beta_scale = nn.Parameter(torch.ones(1, device=device) * 0.01)

        if self.use_hc_norm:
            self.layer_norm = LayerNorm(dim // self.rate)
    
    def width_connection(self, h):
        # get alpha and beta
        h_shape = h.shape
        h_reshape = h.reshape(h_shape[:-1] + (self.rate, h_shape[-1] // self.rate) )
        if self.use_hc_norm:
            norm_h = self.layer_norm(h_reshape)
        else:
            norm_h = h_reshape

        if self.use_tanh:
            dynamic_alpha = F.tanh(norm_h @ self.dynamic_alpha_fn)
        else:
            dynamic_alpha = norm_h @ self.dynamic_alpha_fn
        
        if self.use_scale:
            dynamic_alpha = dynamic_alpha * self.dynamic_alpha_scale

        alpha = dynamic_alpha + self.static_alpha[None, None, ...]

        if self.use_tanh:
            dynamic_beta = F.tanh(norm_h @ self.dynamic_beta_fn)
        else:
            dynamic_beta = norm_h @ self.dynamic_beta_fn

        if self.use_scale:
            dynamic_beta = dynamic_beta * self.dynamic_beta_scale

        beta = dynamic_beta + self.static_beta[None, None, ...]
        mix_h = (alpha.transpose(-1, -2).contiguous().float()  @  h_reshape.float()).bfloat16()
        
        return mix_h, beta
    
    def depth_connection(self, mix_h, h_o, beta):
        h_o_shape = h_o.shape
        h = beta[..., None] * h_o.reshape(h_o_shape[:-1] + (self.rate, h_o_shape[-1]//self.rate)) + mix_h[..., self.rate:, :]
        h_shape = h.shape
        
        return h.reshape(h_shape[:-2] + (h_shape[-2] * h_shape[-1], ))
\end{lstlisting}
\end{algorithm}

\clearpage
\begin{algorithm}[H]
\caption{Pseudocode of transformer with frac-connections in a PyTorch-like style.}
\label{alg:torch_trans_with_fc}
\algcomment{\fontsize{7.2pt}{0em}\selectfont 
%\vspace{-1.em}
}
\definecolor{codeblue}{rgb}{0.25,0.5,0.5}
\lstset{
  backgroundcolor=\color{white},
  basicstyle=\fontsize{7.2pt}{7.2pt}\ttfamily\selectfont,
  columns=fullflexible,
  breaklines=true,
  captionpos=b,
  commentstyle=\fontsize{7.2pt}{7.2pt}\color{codeblue},
  keywordstyle=\fontsize{7.2pt}{7.2pt},
%  frame=tb,
}
\begin{lstlisting}[language=python]
# h: hidden vector (BxLxD)
# atten_frac_connection, ffn_frac_connection:  frac-connection modules
# attn_norm, ffn_norm: normalization modules

# Attention Block
mix_h, beta = atten_frac_connection.width_connection(h)
mix_h_shape = mix_h.shape
h = mix_h[...,:self.rate,:].reshape(mix_h_shape[:-2] + (mix_h_shape[-2] // 2 * mix_h_shape[-1], ))
h = attn_norm(h)
h = self_attention(h)
h = atten_frac_connection.depth_connection(mix_h, dropout(h), beta)

# FFN Block
mix_h, beta = ffn_frac_connection.width_connection(h)
mix_h_shape = mix_h.shape
h = mix_h[...,:self.rate,:].reshape(mix_h_shape[:-2] + (mix_h_shape[-2] // 2 * mix_h_shape[-1], ))
h = ffn_norm(h)
h = ffn(h)
h = ffn_frac_connection.depth_connection(mix_h, dropout(h), beta)

\end{lstlisting}
\end{algorithm}

\clearpage
\section{OLMo2 Model Results}
\label{app:olmo2_result}
\begin{figure}[H]
    \begin{center}
    \includegraphics[width=0.9\textwidth]{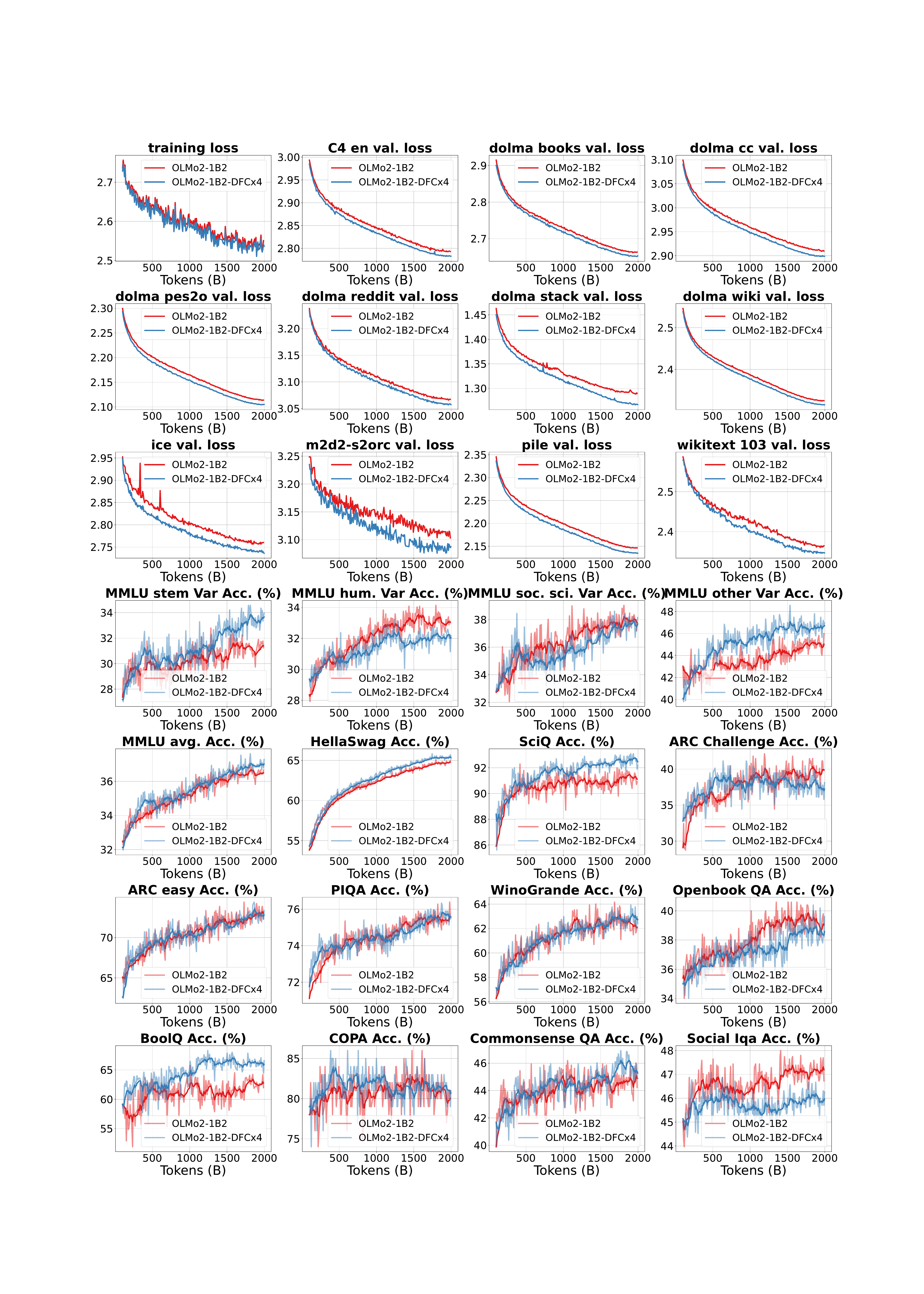}
    \end{center}
  \caption{Loss and accuracy curves for \texttt{OLMo2-1B2} and \texttt{OLMo2-1B2-DFC$\times{4}$} models.}
    \label{fig:moe2_full_results}
\end{figure}

\clearpage
\section{OLMoE-7B Model Results}
\label{app:olmo2_result}
\begin{figure}[H]
    \begin{center}
    \includegraphics[width=0.9\textwidth]{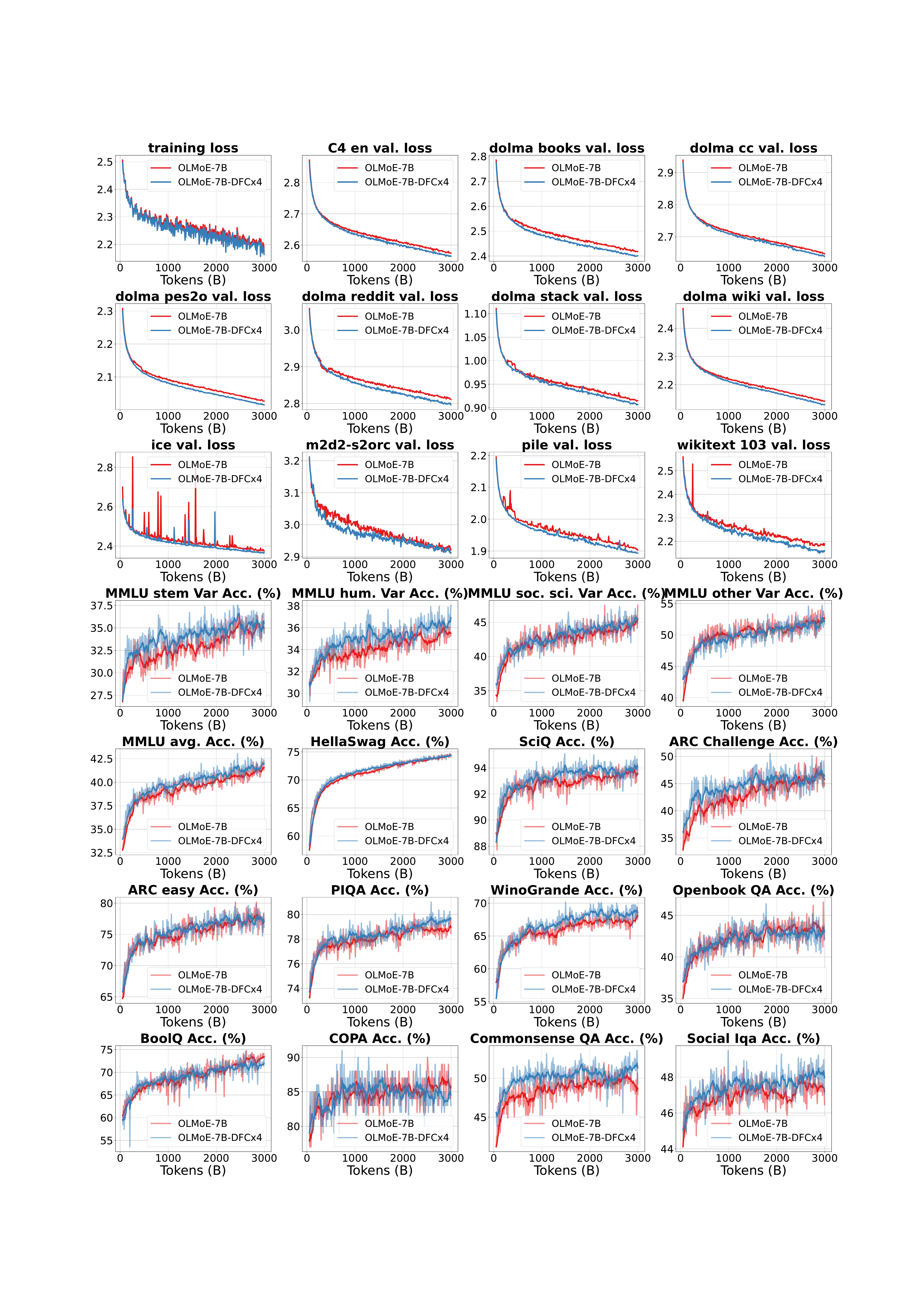}
    \end{center}
  \caption{Loss and accuracy curves for \texttt{OLMoE-7B} and \texttt{OLMoE-7B-DFC$\times{4}$} models.}
    \label{fig:moe_1b7b_full_results}
\end{figure}

\section{Downstream Benchmarks}
\begin{table}[H]
\centering
\caption{Downstream Benchmarks.}
\begin{tabular}{|l|}
\hline
\multicolumn{1}{|c|}{\textbf{Downstream Benchmarks}} \\
\hline
\texttt{piqa}~\citep{bisk2020piqa} \\
\texttt{hellaswag}~\citep{zellers2019hellaswag} \\
\texttt{winogrande}~\citep{sakaguchi2021winogrande} \\
\texttt{openbook\_qa}~\citep{OpenBookQA2018} \\
\texttt{sciq}~\citep{SciQ} \\
\texttt{arc\_easy}~\citep{allenai:arc} \\
\texttt{arc\_challenage}~\citep{allenai:arc} \\
\texttt{copa}~\citep{roemmele2011choice} \\
\texttt{boolq}~\citep{clark2019boolq} \\
\texttt{commonsense\_qa}~\citep{talmor2018commonsenseqa} \\
\texttt{social\_iqa}~\citep{sap2019socialiqa} \\
\texttt{mmlu}~\citep{hendryckstest2021} \\
\hline
\end{tabular}
\label{tab:benchmarks}
\end{table}

\end{document}